\documentclass[10pt,twocolumn,letterpaper]{article}

\usepackage{3dv}
\usepackage{times}
\usepackage{epsfig}
\usepackage{graphicx}
\usepackage{amsmath}
\usepackage{amssymb}
\usepackage{booktabs}
\usepackage{capt-of}

\usepackage[numbers,sort,compress]{natbib} 

\usepackage{mathtools}

\usepackage[pagebackref=true,breaklinks=true,colorlinks,bookmarks=false]{hyperref}

\threedvfinalcopy %

\def\threedvPaperID{242} %

\ifthreedvfinal\fi

\usepackage{newfloat}
\DeclareFloatingEnvironment[
    fileext=los,
    listname={List of appendix figures},
    name=Figure,
    placement=tbhp,
    within=section,
]{apdxfig}

\newcommand{\myparagraph}[1]{\vspace{0.2em}\noindent\textbf{#1}}

\newcommand{\KK}[1]{{\color{black}#1}} %

\newcommand{\methodname}{{HALO~}}

\begin{document}

\title{A Skeleton-Driven Neural Occupancy Representation for Articulated Hands}

\newcommand*{\affaddr}[1]{#1}
\newcommand*{\affmark}[1][*]{\textsuperscript{#1}}
\newcommand*{\email}[1]{\small{\texttt{#1}}}
\author{
Korrawe Karunratanakul \quad
Adrian Spurr \quad
Zicong Fan \quad
Otmar Hilliges \quad
Siyu Tang\\

\affaddr{ETH Z{\"u}rich}\\
\email{\{korrawe.karunratanakul,adrian.spurr,zicong.fan,otmar.hilliges,siyu.tang\}@inf.ethz.ch}
}

\twocolumn[{%
\renewcommand\twocolumn[1][]{#1}%
\maketitle

\begin{center}
  \newcommand{\teaserwidth}{\textwidth}
  \centerline{\includegraphics[width=\linewidth]{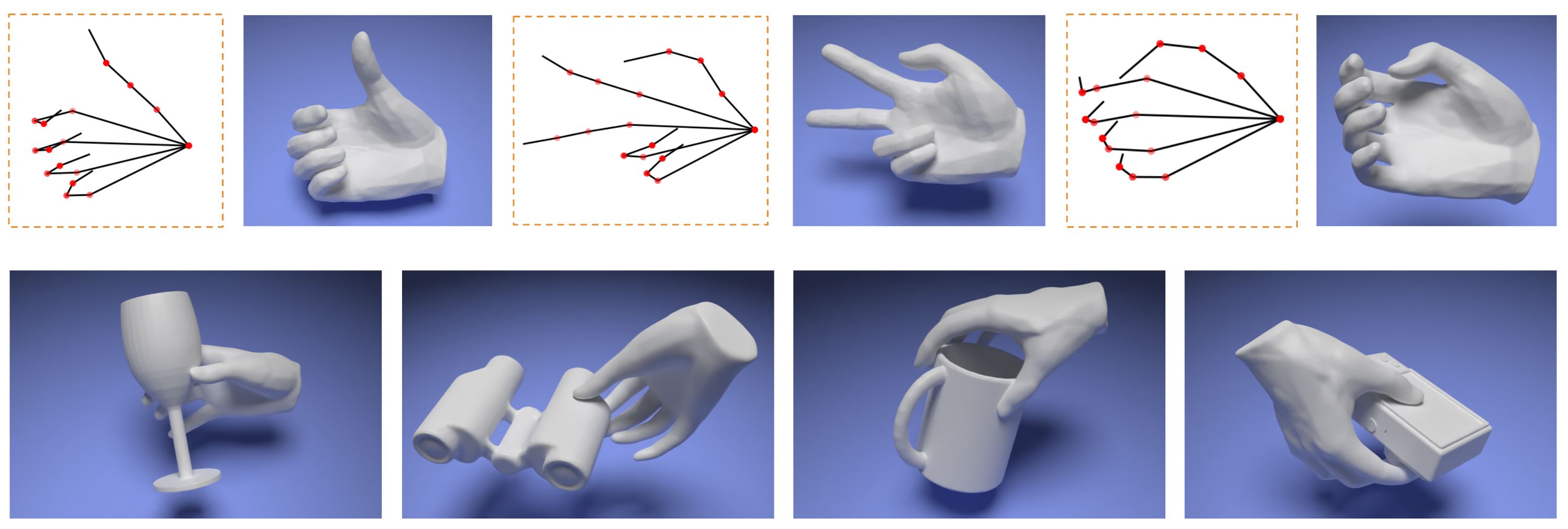}}
    \captionof{figure}{ We introduce a novel neural implicit surface representation of human hands (HALO) that is fully driven by keypoint-based skeleton articulation. Taking 3D keypoints as input, a fully differentiable implicit occupancy representation produces high-fidelity reconstructions of the hand surface (top row). We show that HALO facilitates the conditional generation of articulated hands that grasp 3D objects in a realistic and physically plausible manner (bottom row).}
    \label{fig:teaser}
\end{center}%
}]

\begin{abstract}
We present Hand ArticuLated Occupancy (HALO), a novel representation of articulated hands that bridges the advantages of 3D keypoints and neural implicit surfaces and can be used in end-to-end trainable architectures.  
Unlike existing statistical parametric hand models (e.g.~MANO), HALO directly leverages 3D joint skeleton as input and produces a neural occupancy volume representing the posed hand surface. 
The key benefits of HALO are
(1) it is driven by 3D keypoints, which have benefits in terms of accuracy and are easier to learn for neural networks than the latent hand-model parameters;
(2) it provides a differentiable volumetric occupancy representation of the posed hand;
(3) it can be trained end-to-end, allowing the formulation of losses on the hand surface that benefit the learning of 3D keypoints.   
We demonstrate the applicability of HALO to the task of conditional generation of hands that grasp 3D objects. The differentiable nature of HALO is shown to improve the quality of the synthesized hands both in terms of physical plausibility and user preference. 

\end{abstract}

\section{Introduction}
Humans grasp and manipulate objects with their hands. 
Modeling 3D poses and surfaces of human hands is important for 
numerous applications such as animation, games, augmented and virtual reality. %
Existing hand representations in the literature can be categorized into two paradigms: skeleton representations  ~\cite{ge2016robust,moon2018v2v,Moon_2020_ECCV_InterHand2.6M,iqbal2018hand,zimmermann2017learning} and mesh-based representations~\cite{MANO2017, moon2020deephandmesh}. Even though 3D skeletons are defined in the Euclidean space and are easy to interface with deep neural networks, they lack surface information and therefore not suitable for reasoning about hand-object interaction. In contrast, mesh-based hand models provide surfaces and can thus explicitly reason about physical interactions, such as hand-object manipulation ~\cite{hasson2020leveraging,GRAB2020}. 
However, their pose and shape parametrizations are often hard to directly interpret and more difficult to learn in an end-to-end fashion than 3D keypoints. %

In this work, we aim to bridge the gap between 3D keypoints and dense surface models. To this end, 
we propose {\bf H}and {\bf A}rticu{\bf L}ated {\bf O}ccupancy (HALO), a novel hand representation that is driven by keypoint-based skeleton articulation and provides a high-fidelity, resolution-independent neural implicit surface. 
Importantly, the proposed representation is fully differentiable and can be trained end to end such that volume-based losses can back-propagate gradients to the 3D keypoints. 
Specifically, HALO makes two key innovations: (1) a fully differentiable skeleton canonicalization algorithm and (2) a shape-aware neural articulated hand representation that is driven directly by skeleton and naturally affords differentiable reasoning about surface and volumetric occupancy.

\noindent \textbf{Differentiable skeleton canonicalization.}
The core advantage of the HALO model is that its surface representation is \emph{fully differentiable} and driven by the \emph{3D joint skeleton}.
Even though a naive iterative fitting procedure such as inverse kinematics could provide the transformations needed for changing from the canonical pose to the target skeleton, it prevents gradient flow from the surface to the keypoints. In addition, due to ambiguities of the twist angle around the finger bones, the optimization procedure may lead to unnatural surfaces.
By leveraging bio-mechanical constraints that ensure plausible poses, we propose a novel differentiable layer that converts 3D keypoint locations to the corresponding bone rotations and translations with respect to a canonicalized pose space in a single forward pass. \KK{We call this operation a canonicalization layer.}
Importantly, this layer does not rely on iterative fitting and can therefore be effectively used in back-propagation-based learning.
Furthermore, this layer enables the learning of implicit hand shape representations in the canonical space. This significantly improves the generalization capability of the learned representations across different hand poses and shapes, as shown in Sec. \ref{sec:experiments}.

\noindent \textbf{Skeleton-driven neural articulated hand.}
Recently, several neural occupancy networks for human body modelling, e.g.~\cite{deng2020neural, mihajlovic2021leap, Saito:CVPR:2021}, have been proposed. Despite demonstrating the feasibility of parameterizing articulated deformations via neural implicit surfaces, all these methods require ground-truth bone transformations as input, making them unable to interface with keypoints. 
\KK{In addition, it is uncertain and has not been demonstrated how these models will function under noisy predictions without the ground truth transformations.}
Note that making implicit representations applicable to articulated hands driven by skeleton is non-trivial. The key challenge is how to infer realistic shapes of unseen hands from the skeletons with highly articulated hand poses.
Our solution includes two steps.
First, by leveraging bio-mechanical constraints, we ensure bijective mappings between 3D skeletons and hand surfaces using the canonicalization layer, which greatly simplifies the learning of the implicit surface for highly articulated hand skeletons;  
Second, in the canonical space, we learn both identity-dependent and pose-dependent deformations of the hand surface with a set of multi-layer perceptrons that generalize well across different hand shapes and poses. 
We systematically compare HALO with several baseline methods. 
The results show that HALO significantly improves the accuracy, generality and visual fidelity over the baselines.

\noindent \textbf{Hand grasps synthesis.} 
To demonstrate the utility of the keypoint-driven implicit surface representation, we deploy HALO for conditional generation of hands grabbing 3D objects. 
We propose a novel generation pipeline that synthesizes hand keypoints and yields the neural occupancy of the hand via HALO. 
By exploiting the differentiable nature of HALO, we design a hand-object interpenetration loss to guide the training of the 3D keypoint generator in an end-to-end fashion. 
Our experiments show that this loss leads to convincing hand-object contact of the generated hands. 
Furthermore, compared to a MANO-based method (GrabNet \cite{GRAB2020}), HALO produces more physically plausible and visually convincing grasps even before refinement, suggesting that the volume-based losses are effective for learning the 3D keypoint generator.

Overall, the contributions include 
(1) HALO, the first neural implicit hand model that is driven by keypoint-based skeleton articulation, provides smooth neural implicit surfaces, and enables differentiable reasoning about volumetric occupancy; 
(2) A differentiable skeleton canonicalization layer that maps any skeleton to the canonical pose with a unique bio-mechanically valid transformations; 
(3) A realistic human grasp generation framework that leverages the efficient volumetric occupancy checks enabled by HALO.

\section{Related Work}

\myparagraph{Hand pose and mesh estimation.}
Hand pose estimation is a long standing question and several learning-based approaches have been introduced. These approaches generally involve  predicting 3D key point locations \cite{de2011model, spurr2018cross, mueller2018ganerated, garcia2018first, iqbal2018hand, cai2018weakly, moon2018v2v, tekin2019ho, yang2019disentangling, doosti2020hope, spurr2020eccv, Moon_2020_ECCV_InterHand2.6M, zimmermann2017learning, simon2017hand, wu2005analyzing, ge2016robust}, regressing MANO \cite{MANO2017} parameters \cite{baek2019pushing, boukhayma20193d, hasson2019learning, baek2020weakly, hasson2020leveraging, zhang2019end}, or directly predicting the full dense surface of the hand \cite{ge20193d, Kulon_2020_CVPR, moon2020deephandmesh, wan2020dual}. The methods that directly predict 3D key points usually achieve better performance, however, they do not yield dense surface which is crucial for hand interaction. Iteratively fitting a templated-based model such as MANO to the key points could recover the dense surface but also make the process non-differentiable \cite{panteleris2018using,mueller_siggraph2019,wang_SIGAsia2020}.
Alternatively, the dense surface can also be estimated from 3D or 2D keypoints \cite{zhou2020monocular, Choi_2020_ECCV_Pose2Mesh, yang2020bihand}. However, such estimation could result in a change of hand pose from the input keypoints.
In contrast, our model produce hand surface that faithfully respects the input pose and allows surface or volumetric losses to back-propagate directly to the keypoints.

\myparagraph{Hand representation.}
The surface of 3D hands can be represented explicitly or implicitly.
The most common template-based approaches such as MANO \cite{MANO2017} induce a prior of poses and shapes over its learned parameter space for regularization.
However, using the learned parameters also increases the learning complexity as these features do not correspond directly to features in the inputs such as visible joints.
In \cite{baek2019pushing,boukhayma20193d,hasson2020leveraging,hasson2019learning,zhang2019end}, the MANO parameters are predicted directly using additional weak supervision such as hand masks \cite{baek2019pushing,zhang2019end} or 2D annotations \cite{boukhayma20193d,zhang2019end}.
Another way of representing explicit 3D hand is to directly store the dense vertex locations of the MANO template \cite{ge20193d,Kulon_2020_CVPR,Moon_2020_ECCV_I2L-MeshNet}.
While being more generalizable by avoiding constraints on the parameter space, these approaches require the corresponding, dense 3D annotations, which might be difficult to acquire. 
Our work differs in that we can recover dense hand surfaces from 3D keypoints, eliminating the need for learning model parameters or predicting dense surface points.

\myparagraph{Implicit representation.}
Several works represent object shapes by learning an implicit function using neural networks \cite{chen2018implicit_decoder, deng2020cvxnet, Genova_2020_CVPR, genova2019learning,mescheder2019occupancy, park2019deepsdf,sitzmann2020implicit, mildenhall2020nerf}, which allows for the modeling of arbitrary object topologies with dynamic resolution.
Many approaches for learning such implicit function from various input types were also proposed \cite{liu2019learning,liu2020dist,niemeyer2020differentiable,peng2020convolutional,saito2019pifu,sitzmann2019scene}.
These works focus on rigid objects and do not permit shape deformation.
Recently, the interest is also on learning an articulated implicit function for human body~\cite{deng2020neural, mihajlovic2021leap, wang2021locally, bhatnagar2020loopreg, huang2020arch}. \KK{NASA~\cite{deng2020neural} represents human bodies using a set of implicit functions, but the model is limited to a specific body shape. LEAP~\cite{mihajlovic2021leap} proposes to learn inverse linear blend skinning functions for multiple body shapes, however, it relies on ground truth bone transformation matrices instead of 3D joint locations.
To the best of our knowledge, there are no implicit hand representations that can generalize well to various shapes.}
Grasping Field \cite{karunratanakul2020grasping} learns an implicit function for hand and objects together to represent contact but treats every posed hand as a rigid object. As a result, the complexity of learning a wide range of poses increases significantly. 
\KK{In this work, we leverages bio-mechanical constraints of human hand to learn a novel hand model that only takes a skeleton as input and generalizes to different hand shapes and poses.}

\myparagraph{Hand-object interaction.}
There has been many studies into hand interacting with object in various settings \cite{Brahmbhatt2020contactpose, brahmbhatt2019contactdb, corona2020ganhand, karunratanakul2020grasping, GRAB2020, mousavian20196, feix2015grasp,garcia2018first, hampali2020honnotate, grady2021contactopt, chao_cvpr2021, liu2021semi}.
Recently, the community has begun exploring the task of generating plausible hand grasps given an object with notable studies including \cite{corona2020ganhand}, \cite{karunratanakul2020grasping}, and \cite{GRAB2020}.
GanHand \cite{corona2020ganhand} generates grasps suitable for each object in a given RGB image by predicting a grasp type from grasp taxonomy \cite{feix2015grasp} and its initial orientation, then optimize for a better contact with the object.
GrabNet \cite{GRAB2020} uses Basis Point Set \cite{prokudin2019efficient} to represent 3D objects as input to generate MANO parameters. The predicted hand is then fed to a refinement model to improve the contact.
Grasping Field \cite{karunratanakul2020grasping} learns a signed distance field for both hand surface and object surface in one space, allowing the contact to be learned as regions where distances to both surfaces are zeros. However, the output surface cannot be articulated and requires hand model fitting.
Our work differs from others in the way that we use the proposed hand representation to model the contact while keeping the synthesis task as simple as generating 3D keypoints.

\section{HALO: Hand ArticuLated Occupancy} \label{sec:halo}
The HALO model is a skeleton-driven neural occupancy function, formally defined as $\mathcal{O}_w(x|\boldsymbol{J}) \to \{0,1\}$. 
Parameterized by neural network weights $w$, it maps a 3D point $x$ to its occupancy value given the hand skeleton represented by a set of 3D keypoint locations $\boldsymbol{J}$.
In this section, we first describe how to convert an arbitrary 3D joint skeleton to the reference canonical pose in a differentiable and consistent manner, 
then we introduce our simple yet effective neural occupancy networks for hands.%

\myparagraph{Notations.}
Given a hand skeleton represented by 3D key points $\boldsymbol{J}: \mathbb{R}^{21 \times 3}$, we denote
$\theta^f_{i}$ and $\theta^a_{i}$ to be the flexion/extension and abduction/adduction angles of bone $i$ relative to its parent, respectively. For simplicity, we refer to them as flexion and abduction angle. The angle between a palmar bone $i$ and its adjacent palmar bone $i+1$ is denoted as $\theta^p_{i}$. Lastly, $\theta^n_{i}$ is the palmar plane angle between plane $n_i$ and $n_{i+1}$ spanned by the palmar bone $i-1$, $i$, $i+1$ respectively. We denote the properties of the reference canonical hand with $\prescript{c}{}{(.)}$. 
We refer to Supp.~Mat. for further details.

\begin{figure}
    \centering
    \includegraphics[width=\linewidth]{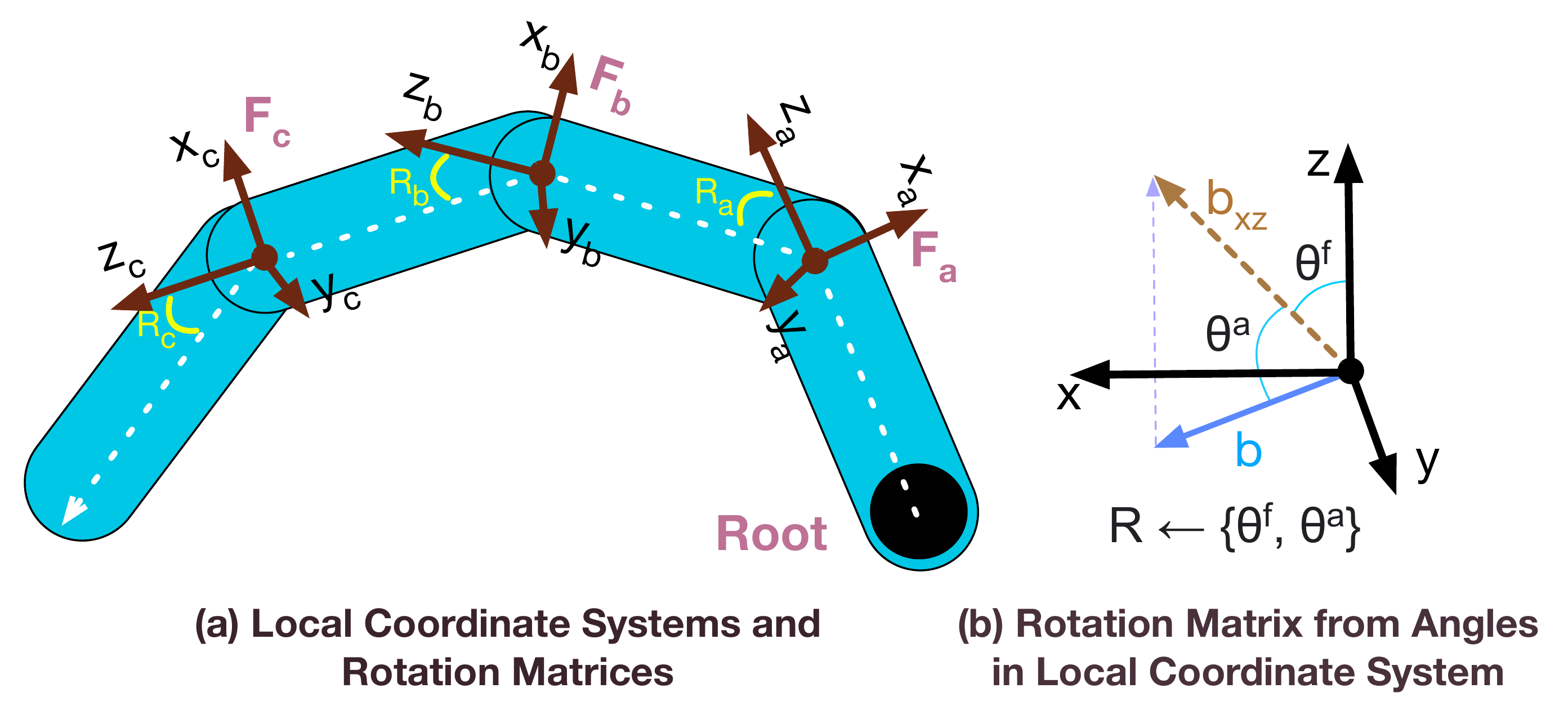}
    \caption{\textbf{Local coordinate systems and rotation matrices defined in the systems.} (a) We adopt the technique from \cite{spurr2020eccv} to construct local coordinate systems $\boldsymbol{F}_i$ for each segment of the kinematic chains of the hand. (b) Each $\boldsymbol{F}_i$ is constructed by measuring the flexion angle $\theta^f$ and the abduction angle $\theta^a$ relative to the parent bone. See Supp. Mat. for further details.}
    \label{fig: angles_coords}
\end{figure}

\subsection{Canonicalization of 3D Hand Skeleton}
Our goal is to learn a neural representation of the surface of human hands in a canonical space. Furthermore, we want to deform this shape based on the spatial configuration of the underlying skeleton, represented by 3D keypoints. 
To do so, we require a mechanism that allows us to convert the 3D keypoints into valid skeletons in the canonical pose in terms of joint angles.
As the keypoints have no notion of the surface, naively converting them to axis-angles does not work due to the unconstrained twist of bones. While twist does not affect keypoints, they affect the surface.

\KK{We take inspiration from Spurr et al.~\cite{spurr2020eccv} which defines a consistent local coordinate system for each bone to measure the bone angles for semi-supervised learning. Our objective is to derive a differentiable mapping layer that i) provides means to convert predicted keypoints to the rest pose and back, and ii) ensures that the skeleton is free of implausible twist that would influence the surface.}

Building on~\cite{spurr2020eccv}, we represent each finger bone by two rotation angles, flexion and abduction, relative to its parent bone (Fig.~\ref{fig: angles_coords}). Each bone cannot rotate about itself, thus, no twist.
However, such formulation ignores the palm configuration, which is needed for defining the canonical pose.
In this work, we propose a method to parameterize the pose of a palm in order to define a consistent canonical pose.
We decouple the palmar bone configuration into 1) finger spreading and 2) palm arching. The spreading of fingers is captured via the angles between two adjacent palmar bones. The arching of the palm is defined by the angle between the two planes spanned by three adjacent palmar bones. The resulting palmar region then serves as a frame of reference for the remaining fingers. 
Please refer to Supp.~Mat.~Fig.~\ref{fig: notation} for better visualization.

\myparagraph{Converting 3D Keypoints to Bone Transformations.}
Formally, we seek the unique set of transformations $\{\boldsymbol{B}^{-1}\}$ that maps the skeleton $ \boldsymbol{J}: \mathbb{R}^{21 \times 3}$ to the canonical pose $\prescript{c}{}{\boldsymbol{J}}$. 
\KK{
Given a skeleton, we obtain the set $\{\boldsymbol{B}^{-1}\}$ by sequentially performing the following operations:
First, we rotate each finger to match the description of our canonical palm pose, which we define as a flat hand with fixed angles between palmar bones;
Second, we compute joint angles and local coordinate systems following \cite{spurr2020eccv} (Fig.~\ref{fig: angles_coords}b), which we use to iteratively undo the angles along the kinematic chain (Fig.~\ref{fig: angles_coords}a) to acquire the canonical pose.
By combining the transformations from both steps, then adjust for the conversion from keypoints to bone vectors, we could obtain a set of transformation matrices $\{\boldsymbol{B}^{-1}\}$ that maps the given skeleton $\boldsymbol{J}_i$ to our canonical pose $ \prescript{c}{}{\boldsymbol{J}_i}$. Formally,
\begin{equation}
    \begin{aligned}
        \prescript{c}{}{\boldsymbol{J}_i} &= \boldsymbol{B_i}^{-1} \boldsymbol{J_i} \quad ,\\
        \text{where } \boldsymbol{B}^{-1} &= \boldsymbol{T S F}' \boldsymbol{R}^c \boldsymbol{F P K} \quad .
    \end{aligned}
    \label{eqn:transMat}
\end{equation}
Here $\boldsymbol{K}: \mathbb{R}^{3 \times 21} \mapsto \mathbb{R}^{3 \times 20}$ is a function that maps the keypoints $\boldsymbol{J} \in \mathbb{R}^{21 \times 3}$ to bone vectors by translating them to the local origin and scaling to unit norm; $\boldsymbol{P}$ normalizes the palmar bone and palmar planes angles; $\boldsymbol{F}$ then maps the bones to their local coordinate frames; $\boldsymbol{R^c}$ rotates each bone to have the same flexion and abduction angles as the canonical pose. Finally, $\boldsymbol{F'}$ maps each coordinate frame back to the global coordinate system; $\boldsymbol{S}$ reverts bones back to their original length and $\boldsymbol{T}$ translates the bones to the tip of their parent bones.

This set of transformations is unique for each skeleton pose and only allows biomechanically valid transformation.
For details, we kindly refer the reader to our Supp.Mat.
}

\begin{figure*}[t!]
    \centering
    \includegraphics[width=0.9\linewidth]{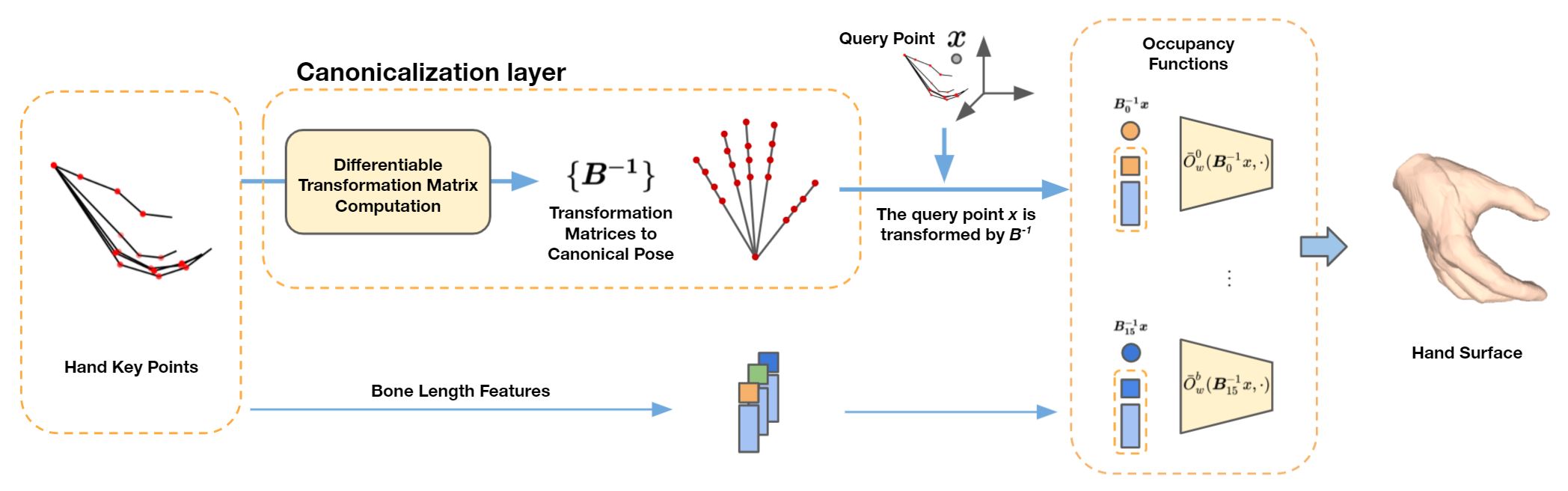}
    \caption{\textbf{Overview of HALO}. 
    Given a hand skeleton, HALO derives bone transformations $\{\boldsymbol{B}^{-1}\}$ that map the bones to the canonical pose using the canonicalization layer. The query point $x$ is then transformed into the canonical space as $\boldsymbol{B_i}^{-1}x$ for the occupancy check.
    }
    \label{fig:halo_model}
\end{figure*}

\begin{figure}[t!]
    \centering
    \includegraphics[width=\linewidth]{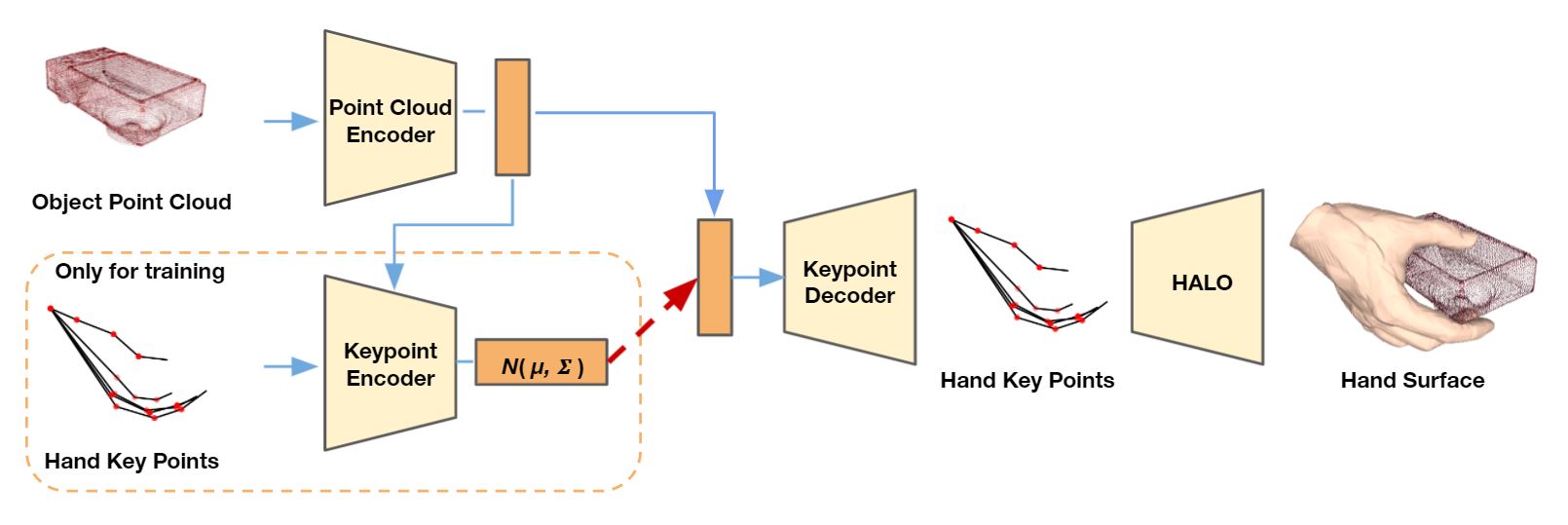}
    \caption{\textbf{HALO-VAE Architecture}. 
    Given object point cloud, the VAE model synthesizes a grasping hand represented by a set of 3D keypoints. The input hand keypoints and the hand encoder are only used during the training.
    From the synthesized hand skeleton, HALO then produces the hand surface. The whole pipeline is end-to-end trainable, therefore, the VAE can leverage volume-based losses to improve the generation quality of the 3D keypoints.
    }
    \label{fig:halo_vae_model}
\end{figure}

\subsection{Neural Occupancy Networks for Hands}
Here, we describe how to leverage the unique mapping $\{\boldsymbol{B}^{-1}\}$ between the posed skeleton and the canonical skeleton to learn the neural hand representation that generalizes to different shapes with highly articulated poses.
We draw inspirations from NASA \cite{deng2020neural} and explore
similar neural network structure due to its simplicity and efficacy.

\myparagraph{NASA~\cite{deng2020neural}. } 
NASA learns an implicit representation of a human body $\mathcal{O}_w(x|\theta) \to \{0,1\}$, conditioned on the pose descriptor $\theta$.
Specifically, it defines the implicit surface for each body part separately. %
Let $\boldsymbol{B}_b^{-1}$ be the transformation to the canonical pose for bone $b$, NASA can be denoted by:
\begin{equation}
    \mathcal{O}_w(x|\theta) = \max_b\{\bar{\mathcal{O}}^b_w(\boldsymbol{B}^{-1}_b x, \mathit{\Pi}^b_w[\{\boldsymbol{B}^{-1} t_0\}])\}.
\end{equation}
where the pose descriptor $\theta$ is defined by a collection of transformation matrices $\{\boldsymbol{B}_b\}_{b=1}^{B}$, and the probability of $\mathcal{O}_w(x|\theta)$ is derived from the maximum occupancy probability across $B$ child occupancy functions $\bar{\mathcal{O}}^b_w(\cdot, \cdot)$, where each represents the body part of the bone $b$. 
For a query point $x$, each child function $\bar{\mathcal{O}}^b_w(\cdot, \cdot)$ maps $x$ to its local coordinate system by the transformation matrix $\boldsymbol{B}^{-1}_b$, so that the local shape of each body part can be learned.
The term $\mathit{\Pi}^b_w[\cdot]$ is used to provide global pose information to each child function.
Essentially, by querying the occupancy value $x$ using $\boldsymbol{B}^{-1}_b x$, the NASA model learns a template shape and the correction based on the global pose with $\mathit{\Pi}^b_w[\cdot]$. Note that, the bone transformation $\{\boldsymbol{B}_b\}_{b=1}^{B}$ is assumed to be given.
For more details, we kindly refer the reader to \cite{deng2020neural}.

\myparagraph{Neural Occupancy Networks for Hands.} 
A naive adaptation of the NASA model for human hand results in erroneous surface reconstructions as shown in Sec.~\ref{sec:eval_nasa} (Fig.~\ref{fig:qualitative}). 
In order to represent hands with highly articulated \textbf{poses} and diverse \textbf{shapes},
we propose to learn the child occupancy functions by conditioning on a shape descriptor $\beta$, effectively learning $\mathcal{O}_w(x|\theta,\beta)$.
We assume that the identify-dependent deformations of the hand are highly correlated to the bones, hence, we propose the use of a collection of bone lengths as the shape descriptor.
In particular, we propose a simple yet effective bone length encoder $f^b(\boldsymbol{D})$ that takes $\boldsymbol{D}=[d_1, d_2, \cdots, d_B]\in \mathbb{R}^B$ the bone length of individual bones as inputs. 
We emphasize that under the proposed formulation, we could learn the hand surface using only the key points $ \boldsymbol{J}$, as the pose descriptor $\boldsymbol{B}^{-1}_b$ is derived from $\boldsymbol{J}$ by Eq.~\ref{eqn:transMat}.
Our final occupancy function is given by:
\begin{equation}
    \mathcal{O}_w(x|\theta,\beta) = \max_b\{\bar{\mathcal{O}}^b_w(\boldsymbol{B}^{-1}_b x, \mathit{\Pi}^b_w[\{\boldsymbol{B}^{-1} t_0\}], f^b(\boldsymbol{D}))\}
    \label{eq: nano}
\end{equation}
where each implicit function $\mathcal{\bar{O}}^b_w$ learns the corresponding part shape based on  
the hand pose descriptor $\mathit{\Pi}^b_w[\{\boldsymbol{B}^{-1} t_0\}]$ and our bone length descriptor $f^b(\boldsymbol{D})$.

\myparagraph{Shape descriptor variations.}
We investigate two versions of bone length encoders $f^b(\boldsymbol{D})$: the local bone encoder $f^b_l(\boldsymbol{D}) = d_b $ and the global encoder $f^b_g(\boldsymbol{D}) = [d_b; MLP(\boldsymbol{D})]$,
where the MLP for the global encoder has two linear layers.
We follow a similar training strategy as \cite{deng2020neural}, for more training details, please refer to the Supp.~Mat.

\subsection{Skeleton-driven Articulated Hand Model}
\KK{
To build a skeleton-driven articulated hand model, we combine the previously described canonicalization layer and the neural hand surface together. 
Specifically, HALO takes the input 3D keypoints to compute bone transformations $\{\boldsymbol{B}^{-1}\}$ for the occupancy networks using the canonicalization layer.
As the canonicalization layer is differentiable, the model can be trained end to end and allows volume-based losses from the surface to back-propagate to the keypoints. 
The overview of HALO is shown in Fig.~\ref{fig:halo_model}.
Note that the bone lengths $\boldsymbol{D}$ can also be computed from the keypoints.
During inference, only 3D keypoints are needed as input to reconstruct hand surface.
}

\section{Human Grasps Generation}
\label{sec:halo-vae}

We show the applicability of the \methodname model in the challenging task of grasps generation. Given an object, we aim to generate diverse grasps with natural and plausible hand-object interaction.
Our grasp generation pipeline consists of two parts: a 3D keypoints generator based on a variational autoencoder (VAE) and the \methodname model for obtaining the hand surface.

\myparagraph{HALO-VAE Architecture.}
The architecture of the HALO-VAE model is illustrated in Fig.~\ref{fig:halo_vae_model}. During training, the object point cloud is first passed to the object encoder, which is a modification of PointNet \cite{pointnet} with residual connection \cite{mescheder2019occupancy}, 
to obtained an object latent code. 
The object latent code is then concatenated to the 3D hand joint location, $\boldsymbol{J} \in \mathbb{R}^{21 \times 3}$, and passed to the VAE encoder.
The decoder reconstructs the 3D hand joint positions conditioned on the hand and object latent representation. 
From the key points, the surface is obtained using HALO through the skeleton canonicalization layer.

The advantages of using \methodname are two-fold. First, we decouple the complexity of learning the pose, represented by the skeleton, from that of learning the surface that corresponds to the pose; 
Second, the implicit model enables fast intersection tests between hand and object, which can be used to efficiently compute an interpenetration loss. Combined with the differentiable skeleton canonicalization layer, the interpenetration loss can be used to improve the keypoints generator in both end-to-end training and post-optimization refinement.

Our grasp generation pipeline is similar to \cite{GRAB2020} and \cite{karunratanakul2020grasping}, but with the following key differences.
First, in \cite{karunratanakul2020grasping}, the output is a rigid implicit surface that cannot be articulated. To obtain an animatable hand for downstream tasks, additional MANO model fitting is required.
Second, in \cite{GRAB2020}, the grasps generator is trained to produce the MANO parameters
which is not directly related to the Euclidean space where the hand and the object live in.
The challenge of interfacing the MANO parameters with deep neural networks is reflected in the GrabNet (CoarseNet) \cite{GRAB2020} results which will be discussed in the experiments section.

\subsection{Learning and Losses}
To train the VAE model, we use the following losses: the KL-divergence loss on the hand latent $Z$, L2 loss on the predicted key points, L1 bone lengths loss, and the bone angle losses.
The bone angle losses are used to provide additional supervisions for learning the hand structure which consists of 1) flexion angles $\theta^f_{i}$ and abduction angles $\theta^a_{i}$, 2) angles between adjacent palmar bones $\theta^p_{i}$ and 3) angles between palmar planes $\theta^n_{i}$.
The bone angles are the same as used in Sec.~\ref{sec:halo}. The losses are defined as L1 angle difference between the prediction and the ground truth.

\myparagraph{Interpenetration loss.} In addition to the losses on the keypoints, we also use the interpenetration loss on the hand surface to avoid collision between hand and object. The key idea is to penalize every points inside the object that is also occupied by the hand. Concretely, for a set of points sampled inside the object $P_o$ and the predicted key points $\boldsymbol{\bar{J}}$, the interpenetration loss is defined as: 

\begin{equation}
    \begin{aligned}
        \mathcal{L}_{in}(\boldsymbol{\bar{J}}) = \sum_{p \in P_o} \mathcal{O}_w(p|\Theta(\boldsymbol{\bar{J}}), f_g(D_{\boldsymbol{\bar{J}}})), \text{where } \mathcal{O}_w(\cdot) > 0.5
    \end{aligned}
    \label{eq:interloss}
\end{equation}
where $D_{\boldsymbol{\bar{J}}}$ is the bone length vector for $\boldsymbol{\bar{J}}$ and $\Theta(\boldsymbol{\bar{J}})$ maps the predicted key points to the \methodname pose vector $\theta$ using the differentiable transformation matrices in Eq.~\ref{eqn:transMat}.

\subsection{Optimization-based Refinement}
To demonstrate that the efficient intersection tests enabled by HALO can be used for optimization, we refine the sampled hands by changing the global translation $t$ to avoid collision with the object. The refinement is run for 10 steps with the interpenetration loss term in Eq.~\ref{eq:interloss}.
The optimization objective is:

\begin{equation}
    \begin{aligned}
         \min_{t}{\mathcal{L}_{in}(\boldsymbol{\bar{J}} + t)}
    \end{aligned}
    \label{eq:optim}
\end{equation}

This simple optimization step aims at refining the contact after the initial prediction of HALO-VAE. It is analogous to the RefineNet in \cite{GRAB2020}, but with an explicit objective to avoid collision instead of being a neural network denoiser.

\section{Experiments}
\label{sec:experiments}
\KK{In this section, we assess our skeleton-driven hand model and the grasp synthesis pipeline.
First, in Sec. \ref{sec:eval_nasa}, we validate the efficacy of HALO as a neural implicit hand model and compare it to the surface baseline \cite{deng2020neural} and keypoints-to-surface baselines \cite{zhou2020monocular, Choi_2020_ECCV_Pose2Mesh}.
Second, we show in Sec. \ref{sec:eval_grasp_synthesis} that HALO can be used effectively in generative tasks which require surface-based reasoning in form of grasp synthesis.
For more experiments, please see supplementary materials.}

\subsection{Neural Hand Model}\label{sec:eval_nasa}
We first evaluate the performance of the proposed implicit surface representation and analyze the effect of the keypoint-to-transformation mapping layer.

\myparagraph{Training data.}
To train our neural occupancy hand model, we utilize MANO \cite{MANO2017} hand meshes. 
Following \cite{deng2020neural}, for each mesh we sample points with two strategies: 1) uniformly sampling in the hand bounding box, 2) sampling on the surface with additional isotropic Gaussian noise. Only the uniformly sampled points are used for evaluation.
The associated occupancy value of each query point is computed by casting a ray from the sampled point and counting the number of intersections along the ray. 
The ground truth bone transformation matrices are computed along the kinematic chain to transform the template MANO hand into the target pose.
The skinning weights are taken from the skinning weights of MANO.
We use the Youtube3D (YT3D) hands dataset \cite{Kulon_2020_CVPR} in all our experiments. The YT3D training set contains 50,175 hand meshes of hundreds of subjects performing a wide variety of tasks in 102 videos. The test set covers 1,525 meshes from 7 videos.

\begin{table}[t!]
\small
\centering
\begin{tabular}{lccc}
\toprule
 & IoU $\uparrow$ & Cham. (mm) $\downarrow$  & Norm. $\uparrow$ \\ \hline

\multicolumn{1}{l|}{NASA \cite{deng2020neural}} & 0.896    &  1.057 & 0.955 \\ \hline
\multicolumn{1}{l|}{NASA+surf.}                 & 0.883    &  1.177  &  0.944 \\
\multicolumn{1}{l|}{NASA+surf.+local b.}        &  0.913   &  0.884  &  0.950      \\
\multicolumn{1}{l|}{HALO (ours)}                   & \bf{0.932}    &  \bf{0.719}  &  \bf{0.959}   \\ \hline
\multicolumn{1}{l|}{HALO keypoints (ours)}        &  0.930  & 0.740 & \bf{0.959}   \\ 
\bottomrule
\end{tabular}
\vspace{1em}
\caption{\textbf{Evaluation on IoU, Chamfer-distance (L1), and normal consistency score (Norm.) between NASA \cite{deng2020neural} and HALO}. All models are trained using groundtruth bone transformations except for \textit{HALO keypoints} where only the 3D key points are given.
\textit{NASA+surf.} indicates a NASA model with resampled surface points for the skinning loss $\mathcal{L}_s$ and \textit{local. b} indicates that a corresponding bone length is given to each occupancy function.
The results show that HALO outperforms NASA \cite{deng2020neural} on IoU and Chamfer-L1 by large margins, suggesting the superior performance of HALO in terms of fidelity and generality.}

\label{table:results}
\end{table}

\begin{table}[t!]
\footnotesize
\centering
\begin{tabular}{lccc}
\hline
  Methods & IOU $\uparrow$ & Cham. (mm) $\downarrow$ &  MPJPE (mm) $\downarrow$ \\
\hline 
Choi et al. \cite{Choi_2020_ECCV_Pose2Mesh} & 0.43 &  4.651 & 14.1 \\
Zhou et al. \cite{zhou2020monocular} & 0.54 &  2.811 & 7.95  \\
HALO keypoints (ours) & \textbf{0.93} & \textbf{0.740}  &  \textbf{0}  \\
\hline
\end{tabular}
\vspace{1em}
\caption{\textbf{Comparison between the estimated, root-aligned surfaces when only 3D keypoints are given as input.}}
\label{tab:eval_kps_mesh}
\vspace{-0.5cm}
\end{table}

\myparagraph{Evaluation metrics.}
For 3D surface reconstruction evaluation, we compute the mean Intersection over Union (IoU), Chamfer-L1 distance, and normal consistency score \cite{mescheder2019occupancy}.

\begin{figure*}[t]
\centering
 \includegraphics[width=0.95\linewidth]{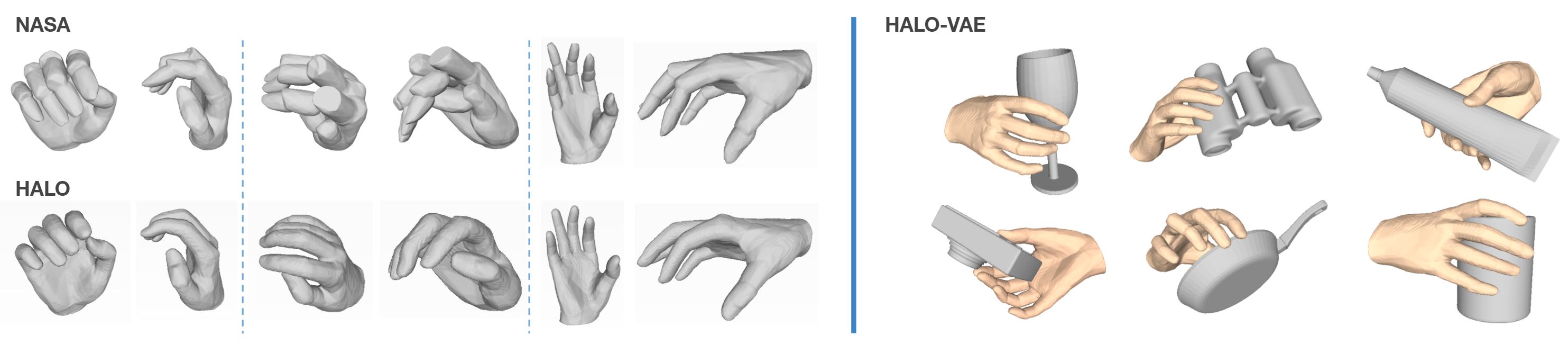}
 \caption{\textbf{Qualitative results of HALO and HALO-VAE.} (Left) Comparison between the NASA and HALO. Two views are shown for the same pose for comparison. (Right) Hands sampled from HALO-VAE.}
 \label{fig:qualitative}
\end{figure*}

\begin{table*}[th]
\footnotesize
\vspace{1em}
\centering
\begin{minipage}{.625\linewidth}

\begin{tabular}{l|ccc|cc}
\toprule
& HALO-VAE               & HALO-VAE              & \cite{GRAB2020}-coarse & HALO-VAE & \cite{GRAB2020}-refine  \\
& w/o $\mathcal{L}_{in}$ & (ours) &                        &  + Optim &  \\
\hline

\multicolumn{1}{l|}{Int.~vol.~(cm3)  $\downarrow$ }  & & & &     \\
\multicolumn{1}{l|}{ / Cont.~ra. (\%) $\uparrow$} & & & &     \\
\multicolumn{1}{l|}{ Binoculars} & 9.19 / 1.00  & \textbf{6.35} / 1.00  & 8.97 / 0.95  & 4.47 / 1.00  & \textbf{3.24} / 1.00 \\
\multicolumn{1}{l|}{ Camera}     & 3.99 / 1.00  & 4.47 / 1.00  & \textbf{3.31} / 0.75  & 1.66 / 1.00  & \textbf{1.46} / 1.00 \\
\multicolumn{1}{l|}{ Frying pan} & 0.25 / \textbf{0.85}  & \textbf{0.22} / 0.65  & 0.94 / 0.80  & \textbf{0.05} / 0.35  & 1.10 / \textbf{0.85} \\
\multicolumn{1}{l|}{ Mug}        & \textbf{3.38} / 1.00  & 6.41 / 1.00  & 5.72 / 1.00  & \textbf{3.48} / 1.00  & 4.48 / 1.00 \\
\multicolumn{1}{l|}{ Toothpaste} & 6.25 / 1.00  & \textbf{1.85} / 1.00  & 6.05 / 1.00  & \textbf{1.11} / 1.00  & 2.28 / 1.00 \\
\multicolumn{1}{l|}{ Wineglass}  & \textbf{1.41} / 0.90  & 2.35 / 1.00  & 2.62 / 1.00  & \textbf{1.56} / 0.80  & 1.60 / \textbf{0.95} \\
\multicolumn{1}{l|}{ Average}    & 4.08 / \textbf{0.96}  & \textbf{3.61} / 0.94  & 4.60 / 0.92  & \textbf{2.06} / 0.85  & 2.36 / \textbf{0.97} \\ \hline

\multicolumn{1}{l|}{Diversity } & & & &      \\
\multicolumn{1}{l|}{ Entropy $\uparrow$}      & \textbf{2.88}  & \textbf{2.88} & 2.84 &  \textbf{2.88} & 2.83 \\
\multicolumn{1}{l|}{ Cluster size $\uparrow$} & \textbf{2.25}  & 2.15 & 1.65 &  \textbf{2.15} & 1.70 \\

\bottomrule
\end{tabular}
\vspace{1em}
\caption{\textbf{Evaluation on interpenetration volume (Int.~vol.), contact ratio (Cont.~ra.) and diversity.} HALO-VAE \small{w/o} $\mathcal{L}_{in}$ denotes the HALO-VAE model {\it without} the interpenetration loss. Best numbers are in bold, except for the contact ratios when more than one model have the same best result.}
\label{table:result_grasp}

\end{minipage} \qquad \qquad
\begin{minipage}{.3\linewidth}

\centering
\footnotesize
\begin{tabular}{cc}
\toprule
\multicolumn{2}{c}{ \% users rated as `more realistic'}   \\ \hline
 HALO-VAE w/o $\mathcal{L}_{in}$  & HALO-VAE \\
 32.77 \%   &  \textbf{67.23} \%            \\ \hline
 HALO-VAE  & GrabNet - coarse  \\
 \textbf{55.75} \%             &  44.25  \%       \\ \hline
 HALO-VAE + Optim  & GrabNet - refine    \\
 \textbf{51.29} \%                      &  48.71 \%              \\ \hline
\bottomrule
\end{tabular}
\vspace{1em}
\caption{\textbf{Binary choice user study.} The number show the percentage of users who rate the corresponding method as more realistic. }
\label{table:userstudy}

\end{minipage}
\end{table*}

\subsubsection{Comparison to implicit surface baseline} \label{sec:transMat_mesh}
Here we investigate the generalization ability of the proposed HALO model to represent articulated hands with various \textbf{poses} and \textbf{shapes}. The results are summarized in Tab.~\ref{table:results}

\myparagraph{Baseline.}
We use the NASA model \cite{deng2020neural} as our baseline. The NASA model is designed to represent an implicit function of an articulated body. However, by changing the input dimension and the number of part-models to match the number of hand parts, it can also be used to represent an articulated hand.
We trained the baseline model using the bone transformation matrices taken from MANO and the sampled query points.
For details on implementation and network architecture we refer to the supplementary.

\myparagraph{Surface vertex re-sampling.}
In \cite{deng2020neural}, the surface vertices $v$ used for enforcing the part models in the skinning loss $\mathcal{L}_s$ are the mesh vertices of SMPL \cite{SMPL2015}. Similarly, we use MANO surface vertices during training.
However, we notice that the human-designed mesh often has many more vertices in the area around the joints which could cause the part models to bias toward the bone endpoints.
Thus, we propose to re-sample the surface vertices uniformly on the mesh surface. 
This result is performance degradation but the bone connections are more natural with less artifact.

\myparagraph{Local and global bone encoders.}
The bone lengths of a human hand greatly influence the hand shape. Therefore, for the local bone encoder, we add the bone length $d_i$ to the back-projected query point as input to the part model $[\boldsymbol{B}^{-1}_i x;d_i]$.
As shown in Tab.~\ref{table:results}, the local bone encoder improves the reconstruction quality both in terms of IoU and Chamfer-L1 distance. We further extend the local bone encoder by considering all the bone lengths as  input. 
A concatenated vector of bone lengths is first fed into a small feed-forward neural network to get the global bone feature $f_g(\boldsymbol{D})$, which is then concatenated with the query point $x$ and the local bone length $d_i$ as input to the part model.

\myparagraph{Results.}
By combining the local and global bone encoders, HALO significantly improves the 3D surface reconstruction quality compared to NASA. 
As shown in Tab.~\ref{table:results}, the IoU is increased from $0.896$ to $0.932$ and the Chamfer-L1 distance is decreased from $1.057mm$ to $0.719mm$.

We provide a qualitative comparison between NASA, and HALO in Fig.~\ref{fig:qualitative}, confirming the quantitative results. The proposed HALO representation generalizes well for highly articulated poses, whereas the NASA model produces severe artifacts at the connection between parts.

\subsubsection{3D keypoints to hand surface} \label{sec:exp_3dkps}
\KK{Tab.~\ref{table:results} also shows the result from HALO that only takes 3D keypoints as input. The keypoint model achieves comparable surface reconstruction performance as when the ground truth transformation are given, showing the effectiveness of our method.
We show the qualitative results in Fig. \ref{fig:teaser} and \ref{fig:qualitative}.

In addition, to evaluate the keypoint-to-surface pipeline, we then compare HALO to the equivalent component in \cite{zhou2020monocular} and \cite{Choi_2020_ECCV_Pose2Mesh} which estimates hand surface from 3D keypoints. The evaluation is done on same the Youtube3D test set where the ground truth 3D keypoints are given as input.
As \cite{Choi_2020_ECCV_Pose2Mesh} requires both 2D and 3D coordinates as inputs, the 2D keypoints is obtain by projecting the 3D keypoints perpendicular to the palm.
For evaluation, we also report the 3D joint error between the predicted hand and the input joints. This metric measures if the input keyoints are faithfully respected by the models. By design, the HALO model does not change the keypoint locations from input to output, thus does not have this error.
The comparison in Tab.~\ref{tab:eval_kps_mesh} shows that \cite{zhou2020monocular} and \cite{Choi_2020_ECCV_Pose2Mesh} change the hand pose and shape in the prediction while HALO faithfully reconstructs the hand surface according to the given keypoints.
}

\subsection{Grasp Synthesis}
\label{sec:eval_grasp_synthesis}

To assess the utility of HALO in downstream tasks we demonstrate our grasp generative model, HALO-VAE.

\myparagraph{Dataset.} We leverage the recently introduced GRAB dataset \cite{Brahmbhatt_2019_CVPR,GRAB2020} and compare our results to GrabNet \cite{GRAB2020}. We compare both to the initial (coarse) predictions of GrabNet \cite{GRAB2020} and the refined results which matches with our own two-stage generation process. 
The test set contains 6 unseen objects. For each object, we fix the object orientation and sample 20 hand proposals from each model.

\myparagraph{Physics Metrics.}
Following \cite{zhang2020generating,zhang2020place}, we evaluate the physical plausibility (interpenetration volume and contact ratio) and diversity, and provide results from a perceptual study.
To evaluate the interpenetration and contact, we measure the ratio of frames in which the hand is in contact with the object and average the interpenetration volume. The volume is calculated by voxelizing hand and object mesh with 1mm cubes and counting the number of intersecting cubes.

\myparagraph{User study.}
We asked 75 participants in a forced-alternative-choice perceptual study to `select the grasp that is more realistic'. For each question, the user is shown 4 views per grasp and forced to select one. We compare all possible combinations on the same object. Each question is assigned to at least 2 participants, totaling 4,800 data points per pair of model comparison. To ensure that the grasps from HALO-VAE and GrabNet have the exact same texture, we fit MANO to our generated key points for rendering.

\myparagraph{Diversity.}
Following \cite{zhang2020generating}, we compute the diversity of the sampled grasps by performing k-means with 20 clusters on all samples, then evaluate the entropy of the cluster assignment and the average cluster size. More diversity results in higher value for both metrics. We use the flatten key point locations of the hands after aligning the root joint and the plane spanned by middle and index palmar bone as features.

\paragraph{Results.}
We first validate the efficacy of the interpenetration loss (Eq.~\ref{eq:interloss}). %
We compare the HALO-VAE models with and without the interpenetration loss.   
The results show that the interpenetration loss helps in: 1) reducing the collision between the objects and the generated hands (Tab.~\ref{table:result_grasp}\KK{, col.1-2}), and 2) largely improves the user preference of the corresponding model (Tab.~\ref{table:userstudy}, \KK{first row}), demonstrating the efficacy of the proposed neural occupancy representation of articulated hand for reasoning about hand-object interaction.

Next, we compare HALO-VAE with GrabNet \cite{GRAB2020}. Both HALO-VAE and GrabNet-coarse are CVAE based generative models and end-to-end trainable, the key difference is that GrabNet-coarse generates MANO model parameters whereas HALO-VAE generates 3D keypoints. 
As shown in Tab.~\ref{table:result_grasp}, HALO-VAE outperforms GrabNet-coarse by a large margin for interpenetration volume and sample diversity. 
Moreover, the HALO-VAE model without interpenetration also compares favorably to GrabNet-coarse, suggesting that the 3D keypoints based representation is well suited to interface with deep neural networks.

Finally, we compare our optimization-based refinement with GrabNet-refine. To the best of our knowledge, the RefineNet is not trained end-to-end with GrabNet-coarse and used for three steps during the inference.
As shown in Tab.~\ref{table:userstudy}, our refined grasps attain a higher user score, suggesting they are more realistic and natural compared to the grasps refined by GrabNet-refine.

\section{Discussion and Conclusion}
In this work, we introduce HALO, a novel surface representation for articulated hands that can generalize to different hand poses and shapes.
We address the issue of the transformation matrix requirement for inferring the 3D occupancy hand by proposing a skeleton canonicalization algorithm that computes valid transformations from 3D keypoints.
The experiments show that our proposed hand model outperforms the baseline and can represent a wide range of hand poses and shapes.
Finally, we demonstrate the HALO can be used to train an end-to-end grasp generator conditioned on an object and produces hand grasps with natural and realistic interaction.
We believe that HALO can be useful in future work attempting to reconstruct the surface of articulated hands directly from images via differentiable rendering and for several downstream tasks that need to perform surface-based computation such as collision detection and response.

\section{Acknowledgement}
We sincerely acknowledge Shaofei Wang and Marko Mihajlovic insightful discussions and help
with baselines.

\newpage

{\small
\bibliographystyle{ieee_fullname}
\bibliography{references}
}

\clearpage

\begingroup
\onecolumn 

\appendix
\begin{center}
\Large{\bf A Skeleton-Driven Neural Occupancy Representation for Articulated Hands\\ **Appendix**}
\end{center}

\counterwithin{table}{section}
\setcounter{page}{1}

\section{More Experimental Analysis}
\subsection{MANO Parameters vs 3D Keypoints}
To demonstrate the applicability of the HALO hand model, we introduce the HALO-VAE model (Sec.~\ref{sec:halo-vae}) for the conditional human grasps generation task. Our model learns to generate the 3D keypoints of a hand that grasps a given object, whereas our baseline model GrabNet~\cite{GRAB2020} generates MANO parameters that represent the grasping hand. As shown in the Sec.~\ref{sec:eval_grasp_synthesis}, the proposed HALO-VAE model largely outperforms the GrabNet in terms of physical plausibility and naturalness of the generated human grasps.

Here, we provide more details and analysis on the object encoding schemes.
The GrabNet~\cite{GRAB2020} encodes the object using the BPS features~\cite{prokudin2019efficient}. Specifically, the BPS encoder is a 4-layers feed-forward network with residual connections between each layer. 
In our trials, we have experimented with this BPS encoder instead of our PointNet encoder. We used the same 4096 basis points as provided in the GRAB dataset. The rest of the architecture is the same to HALO-VAE. 

However, with the near-zero keypoint reconstruction loss on the validation set, 
the generated grasps for a given object are always the same, suggesting that the information from the sampled Gaussian is not used. 
We suspect that during training, the hand keypoints can be inferred using only the object BPS, as the 3D keypoints and the BPS features are highly related. Consequently, the decoder can entirely ignore the features from the hand encoder, producing the same grasp for different samples.

In order to directly compare the keypoint-based and the MANO parameter based grasps generation frameworks, here we use the same object encoding scheme that employs the PointNet architecture~\cite{pointnet}. Specifically, we change the last layer of HALO-VAE (Fig.~\ref{fig:halo_vae_model}) from predicting the hand keypoints ($63$ dimensions, including $3\times 21$ keypoints) to predicting MANO parameters ($61$ dimentions, including 3 global translation, 3 global rotation, 10 shape parameters and 45 pose parameters). Both models are trained without the interpenetration loss.
The results are shown in Tab~\ref{table:kps_mano_comparison}. The keypoint-based generative model produces grasps with better contact and interpenetration while also being more diverse than those generated from the MANO parameters based model, demonstrating the efficacy of the proposed HALO-VAE model.

\begin{table*}[h]
\vspace{1em}
\centering
\begin{tabular}{l|cc}
\toprule
HALO-VAE w/o $\mathcal{L}_{in}$ & Keypoint prediction & MANO parameter prediction \\
\hline

\multicolumn{1}{l|}{Interpenetration volume~(cm3)  $\downarrow$ / Contaxt ratio (\%) $\uparrow$} & &      \\
\multicolumn{1}{l|}{ Binoculars} & 9.19 / \textbf{1.00}  & \textbf{4.88} / 0.95 \\
\multicolumn{1}{l|}{ Camera}     & \textbf{3.99} / 1.00  & 4.00 / 1.00  \\
\multicolumn{1}{l|}{ Frying pan} & 0.25 / \textbf{0.85}  & \textbf{0.21} / 0.50  \\
\multicolumn{1}{l|}{ Mug}        & \textbf{3.38} / 1.00  & 7.22 / 1.00 \\
\multicolumn{1}{l|}{ Toothpaste} & \textbf{6.25} / 1.00  & 6.94 / 1.00  \\
\multicolumn{1}{l|}{ Wineglass}  & \textbf{1.41} / 0.90  & 3.35 / \textbf{1.00}   \\
\multicolumn{1}{l|}{ Average}    & \textbf{4.08} / \textbf{0.96}  & 4.43 / 0.91 \\ \hline

\multicolumn{1}{l|}{Diversity } & &       \\
\multicolumn{1}{l|}{ Entropy $\uparrow$}      & \textbf{2.88}  & 2.85 \\
\multicolumn{1}{l|}{ Cluster size $\uparrow$} & \textbf{2.25}  & 1.13 \\

\bottomrule
\end{tabular}
\vspace{1em}
\caption{\textbf{Comparison between the HALO models predicting key points and predicting MANO parameters.} The results from the keypoint based model are the same as those reported in the main paper.}
\label{table:kps_mano_comparison}
\end{table*}

\begin{apdxfig}
    \centering
    \includegraphics[width=\linewidth]{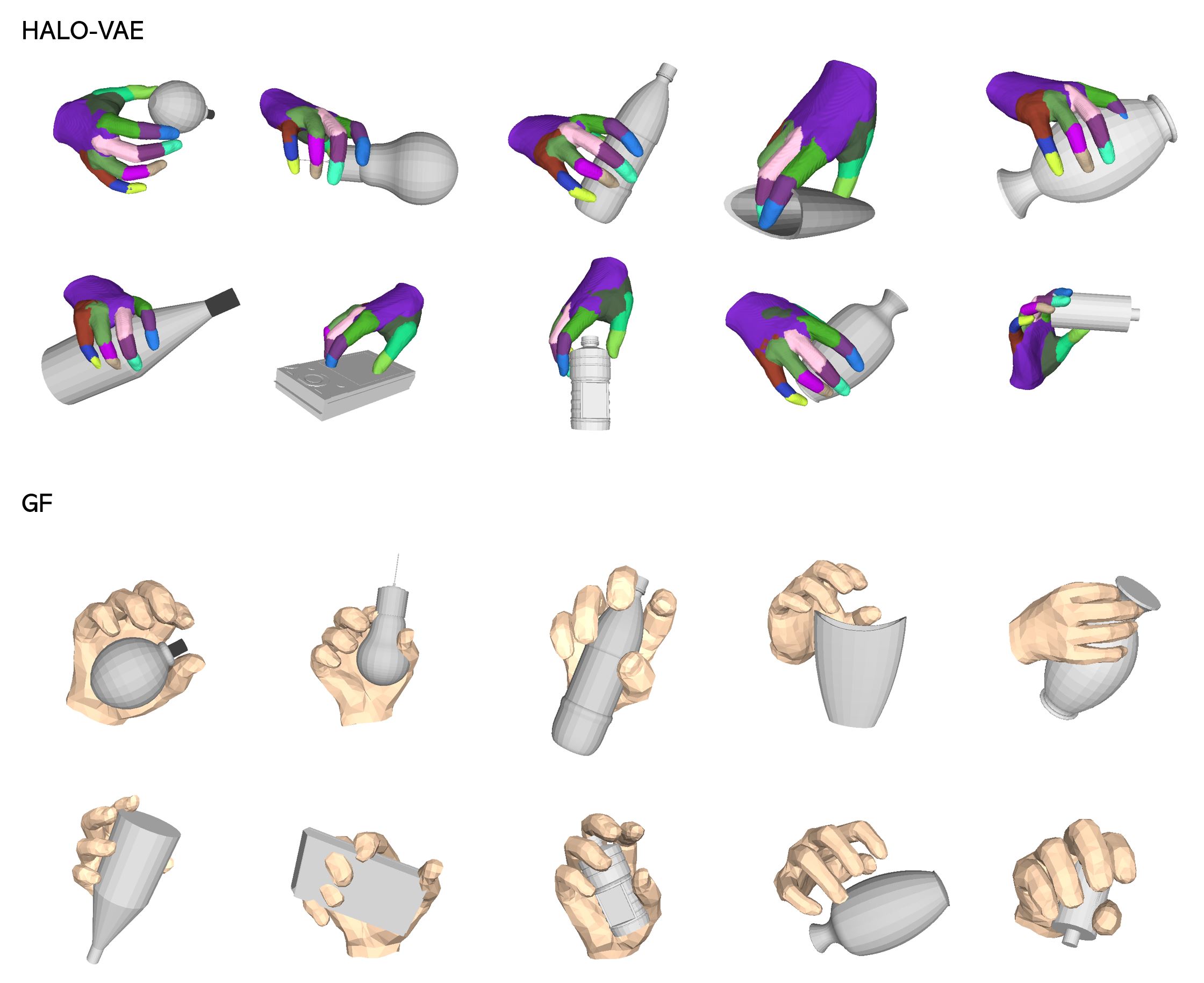}
    \caption{\textbf{Qualitative comparison between HALO-VAE and Grasping Field \cite{karunratanakul2020grasping} on ObMan test objects.} The colors indicate the hand parts based on the child occupancy functions. Our HALO-VAE shows more reasonable grasp and less interpenetration with the object.}
    \label{fig:gf_comparison}
\end{apdxfig}

\subsection{Comparison with Grasping Field \cite{karunratanakul2020grasping} and GrabNet~\cite{GRAB2020} on other datasets}
In this section, we show the comparison between the generated grasps from the Grasping Field (GF) model, Grabnet, and HALO-VAE on the ObMan \cite{hasson2019learning} and HO3D \cite{hampali2020honnotate} test objects.
Note that the HALO-VAE and Grasping Field are not directly comparable as the meshes produced by Grasping Field do not guarantee to be a valid human hand and require MANO fitting, while our HALO-VAE produces articulated implicit hand surfaces.

Nevertheless, we show qualitative and quantitative comparisons between the GF meshes after MANO fitting and the HALO hand surfaces in Fig.~\ref{fig:gf_comparison} and Tab~\ref{table:comparision_obman}, \ref{table:comparision_obman}, respectively.
Due to the artifacts in the hand-designed objects used in the ObMan dataset that interfere with the interpenetration evaluation, e.g non-watertight meshes, surface with holes, internal structure with wrong winding number and zero-volume meshes, we perform the evaluation using the object convex hull instead. 
The evaluation is performed by generating 5 grasps per object on 30 randomly chosen test objects from the ObMan dataset. In total, we evaluate 150 generated grasps from each model.

The results in Tab.~\ref{table:comparision_obman} and \ref{table:comparison_ho3d} show that our HALO-VAE model produces comparable physically-plausible human grasps than Grasping Field~\cite{karunratanakul2020grasping} and GrabNet \cite{GRAB2020} with more diversity.

\begin{table*}[th]
\vspace{1em}
\centering
\begin{tabular}{l|ccc}
\toprule
& HALO-VAE               & Grasping Field & GrabNet - Corase\\
& w/ $\mathcal{L}_{in}$ (ours) &  \\
\hline

\multicolumn{1}{l|}{Interpenetration volume*~(cm3)  $\downarrow$ }  & \textbf{19.95} & 21.93 & 21.82\\
\multicolumn{1}{l|}{Contact ratio*~(\%) $\uparrow$} & 0.98 & 0.90   &  \textbf{1.00} \\ \hline

\multicolumn{1}{l|}{Diversity } & &       \\
\multicolumn{1}{l|}{ Entropy $\uparrow$}      & 2.76           & 2.83 & \textbf{2.88} \\
\multicolumn{1}{l|}{ Cluster size $\uparrow$} & \textbf{3.34}  & 2.87          & 2.68 \\

\bottomrule
\end{tabular}
\vspace{1em}
\caption{\textbf{Comparison with Grasping Field \cite{karunratanakul2020grasping} and GrabNet \cite{GRAB2020} on randomly chosen ObMan test objects.} *Due to the non-watertight object meshes, the interpenetration volume and the contact ratio is approximated using the convex hull of the objects.}
\label{table:comparision_obman}
\end{table*}

\begin{table*}[th]
\vspace{1em}
\centering
\begin{tabular}{l|ccc}
\toprule
& HALO-VAE               & Grasping Field & GrabNet - Corase\\
& w/ $\mathcal{L}_{in}$ (ours) &  \\
\hline

\multicolumn{1}{l|}{Interpenetration volume~(cm3)  $\downarrow$ }  & 25.84 & 93.01 & \textbf{24.62}\\
\multicolumn{1}{l|}{Contact ratio~(\%) $\uparrow$} & 0.97 & \textbf{1.00} & \textbf{1.00} \\ \hline

\multicolumn{1}{l|}{Diversity } & &       \\
\multicolumn{1}{l|}{ Entropy $\uparrow$}      & \textbf{2.81}  & 2.75 & 2.79 \\
\multicolumn{1}{l|}{ Cluster size $\uparrow$} & \textbf{4.87}  & 3.44 & 3.23 \\

\bottomrule
\end{tabular}
\vspace{1em}
\caption{\textbf{Comparison with Grasping Field \cite{karunratanakul2020grasping} and GrabNet \cite{GRAB2020} on HO3D test objects.}}
\label{table:comparison_ho3d}
\end{table*}

\section{Differentiable Bio-mechanical Canonicalization Layer} \label{app:bio_mech}

In this section, we elaborate on the method for converting 3D keypoints to bone transformation matrices. We closely follow the formulation in Spurr \etal \cite{spurr2020eccv} to construct the local coordinate systems $\boldsymbol{F}$. Here we provide a brief summary of the method. For more details on $\boldsymbol{F}$, we refer the readers to the supplementary material of \cite{spurr2020eccv}.

Recall that we seek to compute the set of matrices ${\boldsymbol{B}^{-1}}$, which, in details, is obtained by sequentially performing the following operations: 1) normalizing palmar plane angles, 2) normalizing palmar bone angles, 3) constructing local coordinate frames $\boldsymbol{F}_i$ for each bone with respect to its parent along the kinematic chain, 4) undoing the rotation in the local frames $\boldsymbol{F}_i$, 5) reverting back to the global coordinate frames. Formally,
\begin{equation*}
    \begin{aligned}
        \prescript{c}{}{\boldsymbol{J}_i} &= \boldsymbol{B_i}^{-1} \boldsymbol{J_i} \quad ,\\
        \text{where } \boldsymbol{B}^{-1} &= \boldsymbol{T S F}' \boldsymbol{R}^c \boldsymbol{F P K} \quad .
    \end{aligned}
    \label{eqn:transMatSup}
\end{equation*}
Here $\boldsymbol{K}: \mathbb{R}^{3 \times 21} \mapsto \mathbb{R}^{3 \times 20}$ is a function that maps the keypoints $\boldsymbol{J} \in \mathbb{R}^{21 \times 3}$ to bone vectors by translating them to the local origin and scaling to unit norm; $\boldsymbol{P}$ normalizes the palmar bone and palmar planes angles; $\boldsymbol{F}$ then maps the bones to their local coordinate frames; $\boldsymbol{R^c}$ rotates each bone to have the same flexion and abduction angles as the canonical pose. Finally, $\boldsymbol{F'}$ maps each coordinate frame back to the global coordinate system; $\boldsymbol{S}$ reverts bones back to their original length and $\boldsymbol{T}$ translates the bones to the tip of their parent bones. 

In the following, we define the notations needed and describe the methods for constructing the local coordinate system $\boldsymbol{F}_i$ and each matrix in $\boldsymbol{B}^{-1}$.

\subsection{Notation}

\begin{apdxfig}
    \centering
    \includegraphics[width=0.7\linewidth]{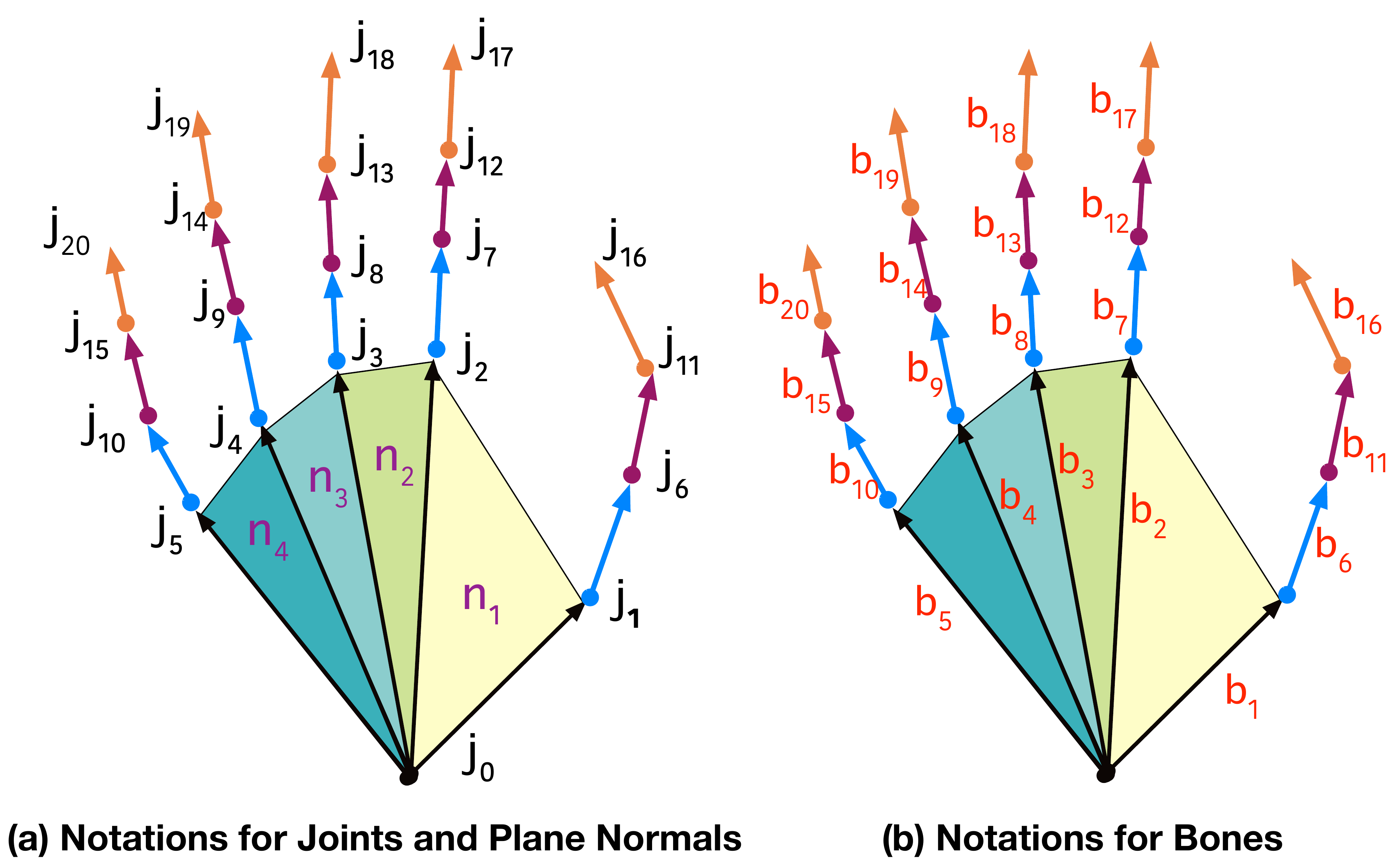}
    \caption{Notations for joints, plane normals and bones for \textbf{right} hand facing up.}
    \label{fig: notation}
\end{apdxfig}

We define all the notations with respect to the right hand. The same procedure could also be applied to the left hand by flipping the x-axis of all joints without loss of generality. 
We denote 3D root-aligned joint locations of a posed hand as $\boldsymbol{J}$ where $j_0$ is the root joint. A bone is defined as a vector pointing from the parent joint to its child $\widetilde{b}_i = j_i - j_{p(i)}$  where $p(i)$ denotes the parent of joint $i$ in the kinematic tree (see Fig.~\ref{fig: notation}). We define $b_i$ as a normalized bone of $\widetilde{b}_i$ and call $\boldsymbol{K}$ the mapping from joints $\boldsymbol{J}$ to normalized bones. 
As a shorthand, we call the palmar bones that are connected to the root joint $j_0$ as the level-0 bones (bones $b_1, \cdots, b_5$). We call a bone with $k$ bone segments in between itself and the root joint a k-level bone.
The bone level from 0 to 3 are denoted by the color black, blue, dark purple, and orange respectively in Fig. \ref{fig: notation}(b).

\textbf{Palmar bone rotations.}
Given a hand skeleton in global coordinate frame, we denote $\theta^p_{i}$ to be the angle between a palmar bone $i$ and its adjacent palmar bone $i+1$; $\theta^n_{i}$ to be the plane angle between plane $n_i$ spanned by the palmar bone $i$, $i+1$ and plane $n_{i+1}$ spanned by the palmar bone $i+1$, $i+2$. We denote the properties of the reference canonical hand with $\prescript{c}{}{(.)}$

\begin{apdxfig}
    \centering
    \includegraphics[width=0.8\linewidth]{figure/angles_coords.pdf}
    \caption{\textbf{(Same as Fig.~\ref{fig: angles_coords}) Local coordinate systems and rotation matrices defined in the systems.} (a) The local coordinate systems $\boldsymbol{F}_a, \boldsymbol{F}_b, \boldsymbol{F}_c$ are constructed based on the kinematic chain. For brevity, we show the kinematic chain of one finger. (b) The flexion angle $\theta^f$ and the abduction angle $\theta^a$ that are constructed based on a bone $b$ in the local coordinate system $\boldsymbol{F}$. The bone $b_{xz}$ is a projection of $b$ in the xz plane.}
    \label{appfig: angles_coords}
\end{apdxfig}

\textbf{Non-palmar bone rotations.}
Given a local coordinate system $\boldsymbol{F}_i = (x_i, y_i, z_i)$ where $(x_i, y_i, z_i)$ are the orientation of the coordinate system, for a bone $b_i$, here we define the flexion $\theta^f_i$ angle and the abduction $\theta^a_i$ angle.
Since each bone $b_i$ is defined the same way in a local coordinate system $\boldsymbol{F}_i$, we drop the subscript $i$ for brevity.  The flexion angle $\theta^f$ of a bone $b$ is the angle between $b_{xz}$ (the projection of $b$ on the $xz$ plane) and the $z$ axis in a coordinate frame $\boldsymbol{F}$. The abduction angle $\theta^a$ is defined by the angle between the bone $b$ and the projection $b_{xz}$ (see Fig. \ref{appfig: angles_coords}(b)).

\subsection{Palmar Bone Normalization ($\boldsymbol{P}$)}
Given a globally normalized hand keypoints, we first compute the transformation matrices $\boldsymbol{P}$ the rotate the palmar bones to match the canonical pose. The palmar bone transformation normalization $\boldsymbol{P}$ is a combination of the palmar plane angle normalization $\boldsymbol{P^p}$ and the palmar bone angle normalization $\boldsymbol{P^a}$, with  $\boldsymbol{P} = \boldsymbol{P^a} \boldsymbol{P^p} $.

\textbf{Palmar Plane Angle ($\boldsymbol{P^p}$)}.
To change the bone angle, we rotate the outer bone (with middle finger being the center) about the shared bone until the plane angle is equal to the canonical angle $ \prescript{c}{}{\theta^p_{i}}$, which we set to 0.8, 0.2, 0.2 radian for $ \prescript{c}{}{\theta^p_{1}}$, $ \prescript{c}{}{\theta^p_{2}}$, $ \prescript{c}{}{\theta^p_{3}}$, respectively.
The plane between the index and middle finger ($n_2$) is fixed as reference.
The rotation applied on the ring finger bone $b_4$ is also propagated to the pinky finger bone $b_5$.

\textbf{Palmar Bone Angle ($\boldsymbol{P^a}$)}.
Secondly, we normalize the spread of the fingers by rotating the bones on the plane two adjacent bones. Concretely, we use the middle finger as reference then rotate $b_2$ on plane $n_2$, $b_1$ on plane $n_1$, $b_4$ on plane $n_3$, and $b_5$ on plane $n_4$. The transformation applied on $b_2$ and $b_4$ are also applied to $b_1$ and $b_5$ respectively.
We set the canonical angle to to 0.4, 0.2, 0.2, 0.2 radian for $ \prescript{c}{}{\theta^n_{1}}$, $ \prescript{c}{}{\theta^n_{2}}$, $\prescript{c}{}{\theta^n_{3}}$, and $\prescript{c}{}{\theta^n_{4}}$ respectively.

\subsection{Local Coordinate System ($\boldsymbol{F}$) }

Now we define the local coordinate system $\boldsymbol{F}_i$ for each bone $b_i$. Note that since in our formulation level-0 bones are always fixed, the only bones that characterize the hand pose are bones at level 1 to 3. Thus, we only describe the coordinate systems for non-zero-level bones.
For each non-zero level bone $b_i$ ($i>5$), its coordinate frame $\boldsymbol{F}_i$ is defined by three normalized vectors $x_i, y_i, z_i$. 
To construct the coordinate system for level-1 to level-3 bones, the z-axis of $\boldsymbol{F}_i$ for non-zero level bones are always defined by the normalized bone vector of its parent $z_i = \text{norm}(b_{p(i)})$. 
We then describe how to define the x-axis. Afterwards, each local coordinate system is defined as $y_i$ can be obtained by a cross product: $y_i = \text{norm}(z_i\times x_i)$. Note that the coordinate frame $\boldsymbol{F}_i$ does not have a position component because we obtain the bone vectors by subtracting the child joint. Thus, all bones are aligned to the origin. The translation components will be added in the final step of our formulation. However, for illustration purposes, we present each coordinate frame $\boldsymbol{F}_i$ with the translation in mind for our figures.

\noindent \textbf{Coordinate Systems for Level-1 Bones} Formally, we denote $n_i$ as the normal of a plane spanned by two adjacent level-0 bones where
\begin{equation}
    n_i = \text{norm}(b_{i+1} \times b_i), \text{ for } i \in \{1,2,3,4\}. 
\end{equation}
For better illustration, Fig. \ref{fig: notation}(a) demonstrates the normal vectors $n_i$ defining each plane. 
Using the normal vectors above, we define the x-axis for level-1 frames $\boldsymbol{F}_i$ for $i \in \{6, 7, 8, 9, 10\}$ as  follows:
\begin{align}
\begin{split}
   x_6 &= -n_1\\
   x_7 &= -n_2\\
   x_8 &= -\text{norm}(n_2 + n_3)\\
   x_9 &= -\text{norm}(n_3 + n_4)\\
   x_{10} &= -n_4
\end{split}
\end{align}

In other words, for bone $b_6$ and $b_{10}$, we define the x-axis for their coordinate system $\boldsymbol{F}_6$ and $\boldsymbol{F}_{10}$ by $-n_1$ and $-n_4$ because they are on the edge of the palm.
For bones $b_8$, and $b_{9}$, the x-axis for their coordinate systems are defined by the average normal around the bones.\\

\noindent \textbf{Coordinate Systems for Level-2 and 3 Bones}  Given the rotation angles $\boldsymbol{R}_{p(i)}$ of the level-1 bones $b_{p(i)} (6 \leq p(i) \leq 10)$ ) and the corresponding coordinate systems $\boldsymbol{F}_{p(i)}$, we can construct coordinate frame $\boldsymbol{F}_{i}$ ($11 \leq i \leq 15$) for the second-level bones by rotating the coordinate frame $\boldsymbol{F}_{p(i)}$ along the kinematic chain using $\boldsymbol{R}_p(i)$.
Concretely, the new coordinate frame is given by
\begin{equation}
    \boldsymbol{F}_{i} = \boldsymbol{R}_{p(i)} \boldsymbol{F}_{p(i)} \label{eq:f}.
\end{equation}
Similarly, the coordinate systems for level-3 bones can be obtained by rotating the coordinate systems of level 2 bones using the rotation angles on level 2.

\noindent\textbf{Rotations to the Canonical Pose ($\boldsymbol{R^c}$)}
Given a local coordinate system $\boldsymbol{F}_i = (x_i, y_i, z_i)$ where $(x_i, y_i, z_i)$ are the axis of the coordinate system, for a bone $b_i$, we can measure the flexion angle $\theta^f_i$ and the abduction angle $\theta^a_i$ that characterize bone $i$ with respect to $\boldsymbol{F}_i$. Fig. \ref{fig: angles_coords} visualizes the local coordinate system $\boldsymbol{F}_i$ and how rotation angles are measured.
Then, given a canonical pose bone $b_i^c$, we compute the rotation matrix $\boldsymbol{R}^c_i$ to transform $b_i$ to its canonical pose $b_i^c$ based on the angles relative to the canonical pose $(\bar{\theta}^f_{i}$, $\bar{\theta}^a_{i})$ in $\boldsymbol{F}_i$.

\subsection{Angles to the Canonical Pose ($\bar{\theta}^f_{i}$, $\bar{\theta}^a_{i}$). ($\boldsymbol{R^c}$)}
Since the angles are measured in a consistent coordinate frame, to obtain the angle needed to rotate to the canonical pose,
we simply offset the angles of $b_i$ by the angle of $b_i^c$ in the canonical pose:
\begin{equation}
    \begin{aligned}
        \bar{\theta}^f_{i} &:= -\theta^f_{i} + \theta^f_{i_c} \\
        \bar{\theta}^a_{i} &:= -\theta^a_{i} + \theta^a_{i_c}  
    \end{aligned} \label{eq:angle2}
\end{equation}
where $\theta^f_{i_c}$ and $\theta^a_{i_c}$ are the flexion and abduction angle of the canonical pose with respect to $\boldsymbol{F}_i$. In our experiments, we use the canonical pose identical to that of MANO \cite{MANO2017}.

\subsection{Local Coordinate Frame to Global Coordinate Frame($\boldsymbol{F'}$).}
Each rotation matrix $\boldsymbol{R}^c_i$ is defined locally with respect to a coordinate frame $\boldsymbol{F}_i$. The accumulated rotations $\boldsymbol{G}_i$ with respect to the root of the hand for a bone $b_i$ is then the product of the rotation matrices along the kinematic chain:
\begin{equation}
\boldsymbol{\boldsymbol{G}_{i}} =
    \begin{cases}
         \boldsymbol{G}_{p(i)} \boldsymbol{R}^c_{i} &\text{if } b_i \text{ is not a root bone}\\ %
         \boldsymbol{I}(4) &\text{otherwise.}
    \end{cases}
\end{equation}
Where $\boldsymbol{G}_{p(i)}$ are the rotations up to the parent of $b_i$. 
With the accumulated rotation matrices encoding the joint-angles relative to the palm of the hand, we need to map all matrices to global coordinates by multiplying with $\boldsymbol{F}_i^{-1}$. Recall that $\boldsymbol{F}_i$ encodes the mapping from the global coordinate system to the local frame. Thus, its inverse brings the angles back to the global coordinate frame. We summarize the accumulation of angles and the mapping to the global coordinate with the matrix $\boldsymbol{F}_i'$:

\begin{equation}
    \begin{aligned}
\boldsymbol{F}_i'      &=   \boldsymbol{F}_i^{-1} \boldsymbol{G}_{p(i)}.
    \end{aligned}
\end{equation}

With all the necessary components in place, we can compute the transformation matrices $\boldsymbol{B}^{-1}$ for all bones using only the 3D keypoints $\boldsymbol{J}$ in a differentiable manner. In Sec. \ref{sec:exp_3dkps} we show that the matrices derived using this formulation can be uses to recover hand surfaces that are very close to those attained by the ground truth transformation matrices from MANO.

\section{Implementation details}
\label{app:implementation}

\subsection{HALO}
\paragraph{Training loss.}
The loss used to train our articulated hand model can be written as:
\begin{equation}
    \mathcal{L} = \mathcal{L}_o + \lambda_s \mathcal{L}_s
\end{equation}
where $\lambda$ determines the weight of the skinning loss. It is set to 0.5 in all experiments. We turn off the skinning loss after the validation IoU reach 80\% as we observe it allows smoother transition between hand parts.

\myparagraph{Network architecture.}
For a fair comparison with the baseline \cite{deng2020neural}, we use the same network architecture with 4 layers of size 40 for each part model in the ablation study.
For the final model used in HALO-VAE, the size is increased to 64 as we observed a small surface quality improvement.
The LeakyRelu with factor 0.1 is used as the activation function. All layers have a residual connection and a dropout of 0.2.
The subspace projection layers $\mathit{\Pi}$ map the input of size $B\times 3 $ to a vector of size 8.
When used, the global bone length encoder is a 2-layer feed-forward network of size 40 that maps 16 bone lengths to an encoded vector of size 16.

We define the number of parts as $B=16$, with one part responsible for the palm and three parts for each finger.
When using the 20 transformation matrices obtained from our formulation as our pose descriptor, we disregard the transformations of the root bones, and add an identity transformation for the palm, resulting in 16 transformation matrices.

\myparagraph{Training data.}
To train our neural occupancy hand model, we utilize MANO \cite{MANO2017} hand meshes. %
The query points for each mesh are selected using two strategies: 1) uniformly sampling points in the bounding box of the hand mesh where the root joint is at the origin, 2) sampling on the surface with additional isotropic Gaussian noise. For each strategy, we sample 100,000 points.
The associated occupancy value of each query point is computed by casting a ray from the sampled point and counting the number of intersections along the ray. 
For evaluation, following~\cite{deng2020neural}, we use uniformly sampled points.
The bone transformation matrices %
are computed along the kinematic chain to transform the template hand into the target pose.
The shape descriptor $\beta$ is based on bone length, defined as the Euclidean distance between adjacent joints.
The skinning weights are taken from the skinning weights for posing the template mesh in MANO.
We use the Youtube3D (YT3D) hands dataset \cite{Kulon_2020_CVPR} in all our experiments. The YT3D training set contains 50,175 hand meshes of hundreds of subjects performing a wide variety of tasks in 102 videos. The test set covers 1,525 meshes from 7 videos.

\myparagraph{Training.}
We used the Adam optimizer with a learning rate $1e-4$ and a batch size of 64 in all experiments.
For each mesh at each training step, we sample 2048 points from the 200K pre-sampled query points for the occupancy loss and 2,000 out of 6,000 surface points for the skinning loss.
When surface point re-sampling is not used, we sampled 200 out of 778 mesh vertices for the skinning loss.

\subsection{HALO-VAE}

\paragraph{Pre-processing.}
For training, we use GRAB dataset \cite{GRAB2020, Brahmbhatt_2019_CVPR} with the default train/test split. The objects are centered at the origin and 600 points are sampled from the surface. The keypoints are obtained by projecting the surface point using MANO.

\paragraph{Network architecture.}
Our keypoint VAE consists of an object encoder, keypoint encoder, and a decoder. The object encoder is a 4-layers PointNet encoder with a residual connection between each layer.
The hand encoder and the decoder are 4-layers MLP networks with residuals connections. The hand encoder that takes hand key points and the object latent code then produces mean and standard variation of a 32-dimension Gaussian distribution. The decoder takes as inputs noise sample from the Gaussian distribution and the object latent vector to predict the hand key point locations. 
All layers have size 256.
From key points to hand mesh, we use the final HALO model with differentiable canonicalization layer that takes 3D key points as inputs.

\section{Limitation }
HALO relies on biomechanically plausible 3D keypoints. 
Training HALO with the model end-to-end with angle losses alleviate this problem and results in a more robust surface prediction, as highlighted in HALO-VAE. This indicates that the inductive bias of our model helps encourage a biomechanically plausible 3D hand surface.
However, a severely physically implausible hand skeleton could still produce artifacts on the hand surface.

\paragraph{Tolerance to Noisy Keypoints.}
We further analyse the impact of the biomechanical violations on the reconstruction quality of hand surfaces. We uniformly sample noises with the amplitude $[-x, +x]$ and add them to every dimension of every joints of a \textit{valid} hand. As shown in Fig.~\ref{fig:noisy_kps}, with the amplitudes are within $2mm$, HALO produces reasonable hand surface. When $x$ is increased to $5mm$, the reconstructed hand surface starts to show visible artifacts.
Nevertheless, we reiterate that this problem can be mitigated by encouraging a bio-mechanically valid skeleton output from the estimator or generator, as in HALO-VAE.

\begin{apdxfig}
    \centering
    \includegraphics[width=\linewidth]{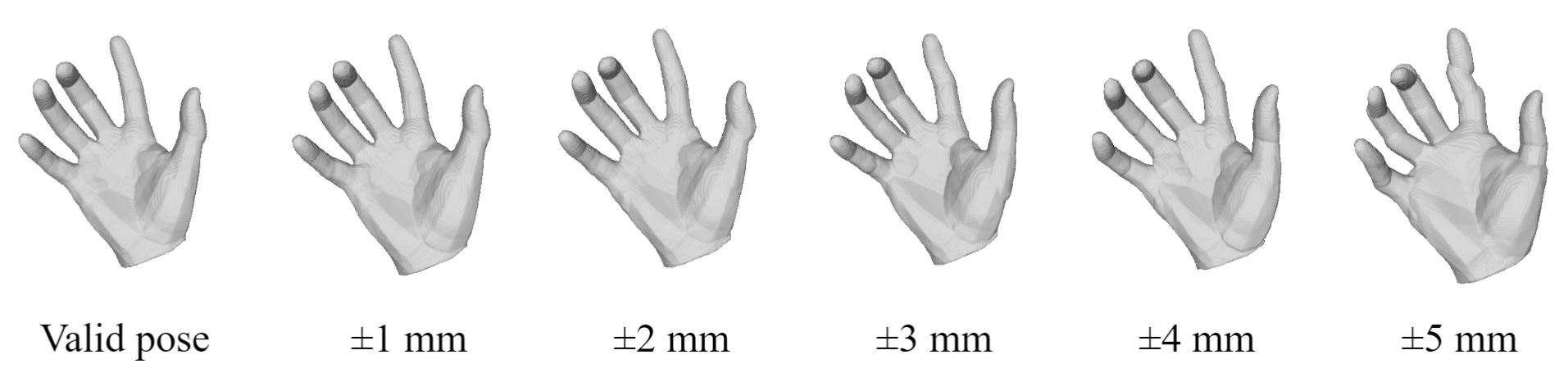}
    \caption{HALO from noisy 3D keypoints.}
    \label{fig:noisy_kps}
\end{apdxfig}

\section{Qualitative Results}
\label{app:qualitative}

\subsection{HALO from Keypoints}
Figure \ref{fig:halo_hand} shows the HALO hand surfaces driven by the keypoint-based skeleton articulations. 

\begin{apdxfig}
    \centering
    \includegraphics[width=\linewidth]{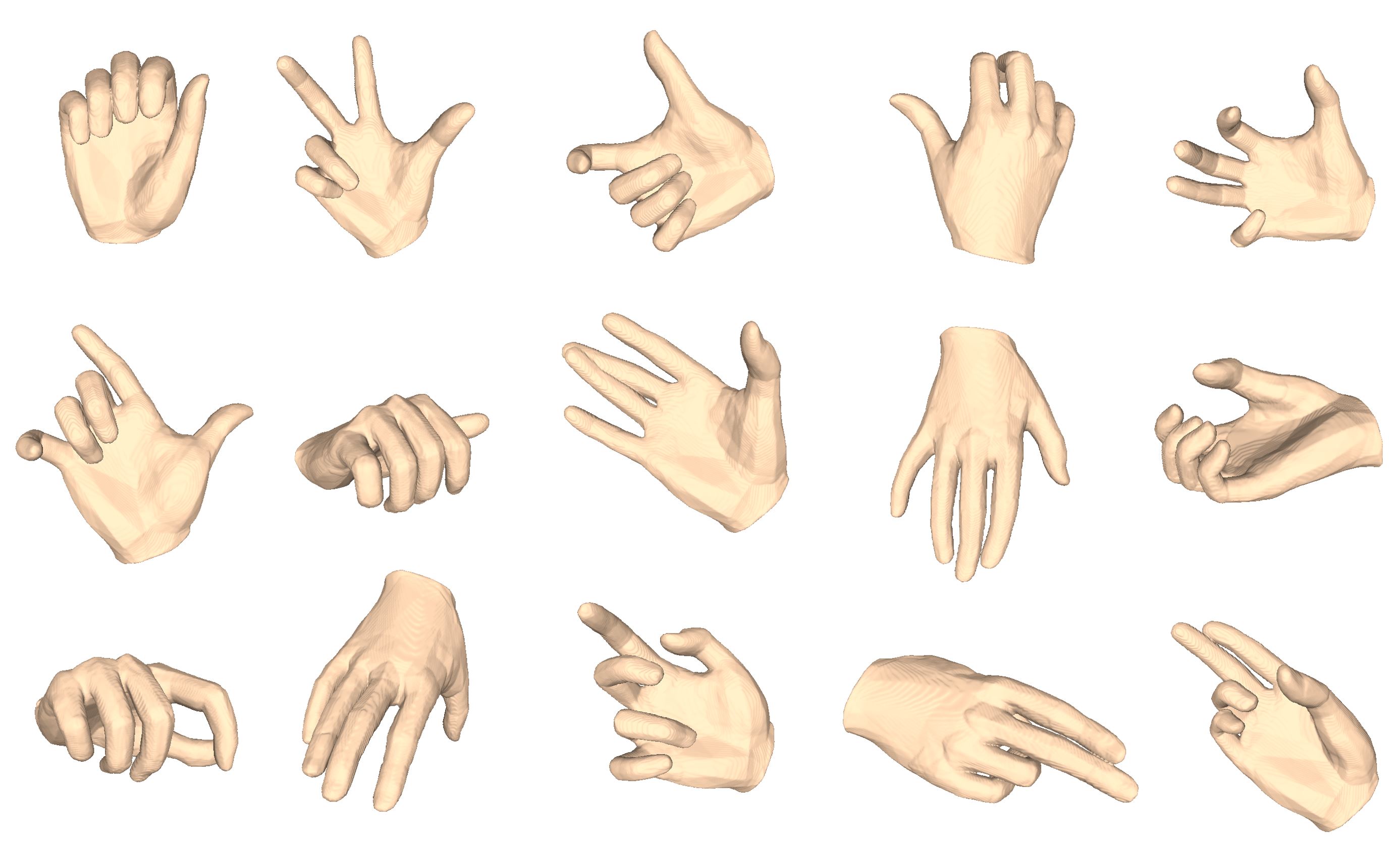}
    \caption{Visualization of the HALO hand surfaces driven by the keypoint-based skeleton articulations. 
    Note that the hand poses (articulated skeletons) are {\bf randomly} sampled from the test set of Youtube3D.}
    \label{fig:halo_hand}
\end{apdxfig}

In addition, to further demonstrate the generalisability of HALO, we show HALO surface driven by ground truth skeletons from the unseen Interhand2.6M~\cite{Moon_2020_ECCV_InterHand2.6M} dataset in Figure~\ref{fig:interhand}.

\begin{apdxfig}[ht]
    \centering
    \includegraphics[width=0.8\linewidth]{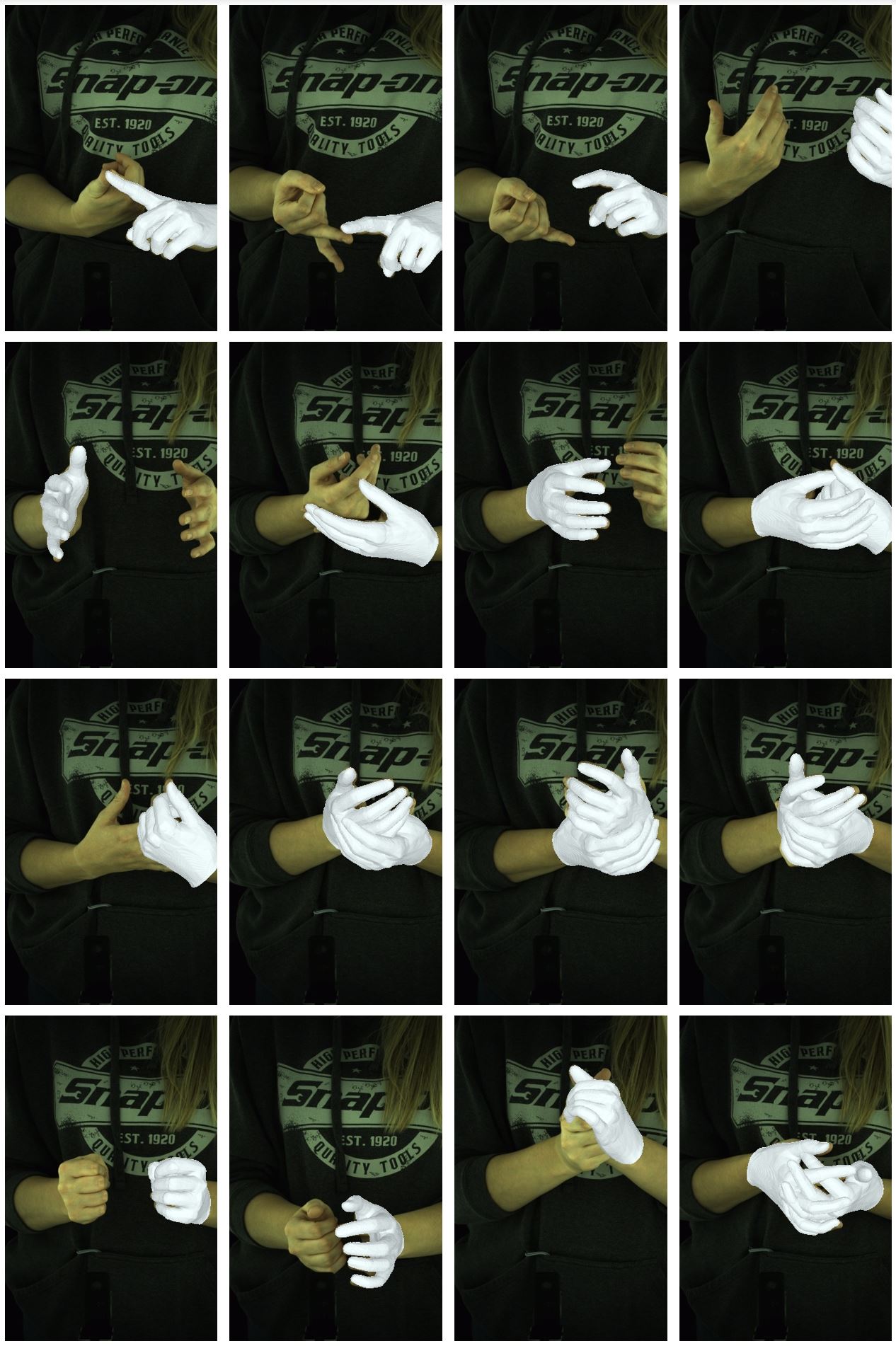}
    \caption{Visualization of the HALO hand surfaces driven by the keypoint-based skeleton from the unseen Interhand2.6M dataset. The ground truth hand keypoints are given as input to reconstruct the hand surfaces. The RGB images are for visualization purpose only, and the results are not from pose estimation.}
    \label{fig:interhand}
\end{apdxfig}

\subsection{Generative Results}

Figure~\ref{fig:halo_vae} shows the grasps {\bf randomly} sampled from HALO-VAE conditioned on the objects from the test set of the GRAB dataset.

\begin{apdxfig}[ht]
    \centering
    \includegraphics[width=0.93\linewidth]{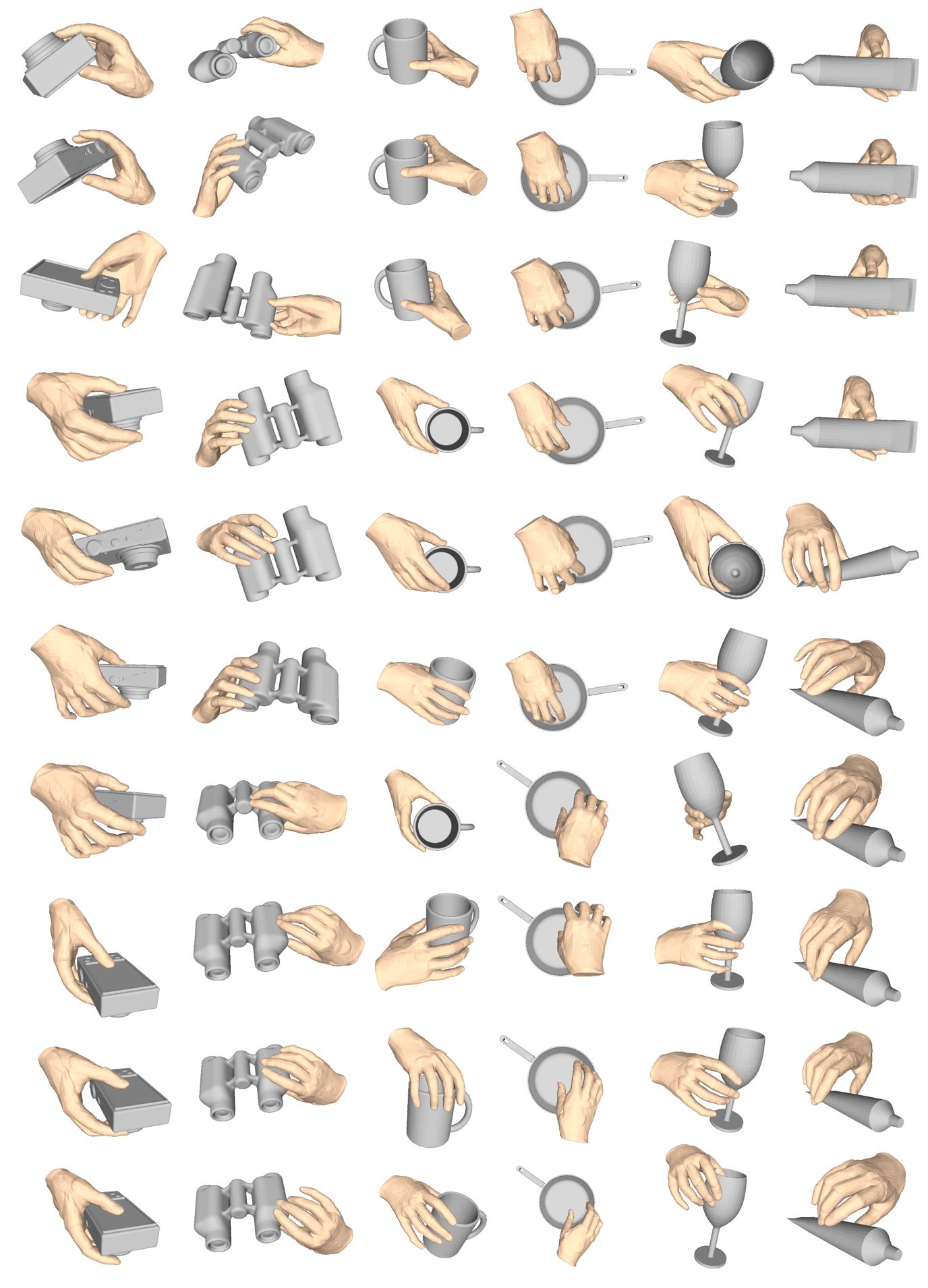}
    \caption{Visualization of the grasps  {\bf randomly} sampled from HALO-VAE conditioned on the 6 unseen test objects from the GRAB dataset, 10 grasps per object.}
    \label{fig:halo_vae}
\end{apdxfig}

\end{document}